\documentclass[journal]{IEEEtran}

\usepackage{amsmath}
\usepackage{xcolor}
\usepackage{bbold}
\usepackage{graphicx}
\usepackage{makecell}
\usepackage{multirow}
\usepackage{booktabs}
\usepackage{tabu}
\usepackage{subfig}

\usepackage[ruled,vlined]{algorithm2e}
\newcommand{\argmin}{\arg\!\min}

\begin{document}

\title{A Real-Time Energy and Cost Efficient Vehicle Route Assignment Neural Recommender System}

\author{Ayman~Moawad,
        Zhijian~Li,
        Ines~Pancorbo,
        Krishna~Murthy~Gurumurthy,
        Vincent~Freyermuth,
        Ehsan~Islam,
        Ram~Vijayagopal,
        Monique~Stinson,
        and~Aymeric~Rousseau
\thanks{A. Moawad is with the Vehicle and Mobility Simulation group at Argonne National Laboratory, 9700 S. Cass Ave, Lemont, IL 60439 USA and the department of Statistics, The University of Chicago, 5801 S. Ellis Ave, Chicago, IL 60637 USA. E-mail: amoawad@anl.gov, aymoawad@uchicago.edu}
\thanks{Krishna Murthy Gurumurthy, Vincent Freyermuth, Ehsan Islam, Ram Vijayagopal, Monique Stinson, and Aymeric Rousseau are with the Vehicle and Mobility Simulation group at Argonne National Laboratory.}
\thanks{Zhijian Li is with the department of Mathematics at The University of California Irvine.}
\thanks{Ines Pancorbo is with the department of Mathematics and Statistics at Georgetown University.}}

\maketitle

\begin{abstract}
This paper presents a neural network recommender system algorithm for assigning vehicles to routes based on energy and cost criteria. In this work, we applied this new approach to efficiently identify the most cost-effective medium and heavy duty truck (MDHDT) powertrain technology, from a total cost of ownership (TCO) perspective, for given trips. We employ a machine learning based approach to efficiently estimate the energy consumption of various candidate vehicles over given routes, defined as sequences of links (road segments), with little information known about internal dynamics, i.e using high level macroscopic route information.
A complete recommendation logic is then developed to allow for real-time optimum assignment for each route, subject to the operational constraints of the fleet. We show how this framework can be used to (1) efficiently provide a single trip recommendation with a top-$k$ vehicles star ranking system, and (2) engage in more general assignment problems where $n$ vehicles need to be deployed over $m \leq n$ trips. This new assignment system has been deployed and integrated into the POLARIS\footnote{POLARIS is an Argonne-based high-performance, open-source agent-based modeling framework for simulating large-scale transportation systems.} Transportation System Simulation Tool for use in research conducted by the Department of Energy's Systems and Modeling for Accelerated Research in Transportation (SMART) Mobility Consortium \cite{smart}.

\end{abstract}


\IEEEpeerreviewmaketitle

\section{Introduction}
\IEEEPARstart{F}{reight} companies are facing increasing pressure to decarbonize their fleets. The large number of technology options, the diversity in vehicle usage and economic uncertainties are major hurdles slowing down new vehicle technology adoptions.
Transportation decarbonization across the freight industry is a major challenge for multiple reasons: (1) the large number of vehicle applications, powertrain and component technologies makes it difficult for fleets to decide which vehicles to invest in, and (2) the diverse vehicle usage, both current and future, raises questions as to which technology should be assigned to a particular route. Understanding the techno-economic impact of technologies is an active field of research. 
Traditionally, energy consumption and economic impacts have been evaluated using standard drive cycles as a baseline for regulatory purposes \cite{moawad_assessment_2016}, \cite{islam_extensive_2018}, \cite{islam_energy_2020}, \cite{islam_energy_2021}, \cite{vijayagopal_fuel_2020}. While they are a good standardization for energy benefit studies,  regulatory cycles fail to represent real driving conditions, traffic variability and other effects such as future infrastructure and connectivity changes, population density changes, traffic behavior changes, etc. 


As a result, the freight industry has been operating under a high level of uncertainty both for longer term technology adoptions and investment as well as day-to-day operations of their current vehicle fleet. Since the industry is heavily driven by cost, we choose to study the optimal assignment problem from a TCO standpoint, assuming that the lowest-cost powertrain option captures driving conditions and number of miles driven over the lifetime of the vehicle under that metric.  

To represent diverse vehicle usage as well as the impact of multiple vehicle technologies on energy consumption and cost, both current and future, we leverage an agent-based transportation tools to model truck routes across the city of Chicago and its suburbs under various scenarios, combined a high-fidelity vehicle system simulation tool, to estimate individual truck energy consumption and cost. The results generated serve as a very large backbone dataset of vehicle-route energy outcomes that capture variability in vehicle classes, powertrain fleet distribution, vehicle technology, automation and connectivity levels, population, driving modes, ride-sharing extent, e-commerce impact, etc. All of these affect traffic and driving behavior: For example, connected and automated vehicle (CAV) technologies are likely to have significant effects not only on how vehicles operate in the transportation system but on how individuals behave and use their vehicles. In Section \ref{sec:DG} we provide high-level details about the data generation process and a brief overview of the data content. For more details about the design of the experiment, the tools involved, and their capabilities, refer to \cite{freyermuth_energy_2019} and \cite{freyermuth_powertrain_2020}.

\begin{figure}[ht!]
\centering
\includegraphics[width=3.4in, clip]{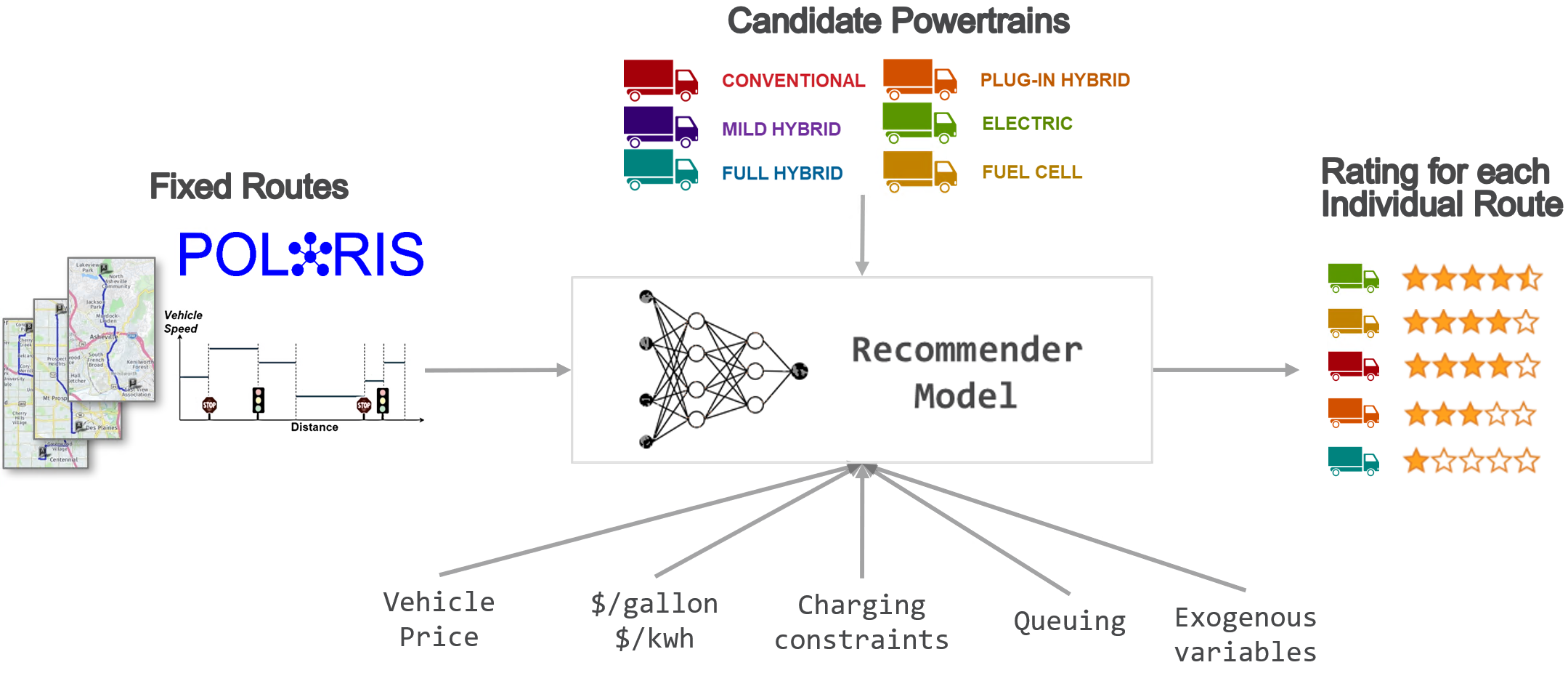}
\caption{Summary view of the recommender system with the required inputs, assumptions and constraints to output a top-\textit{k} ranking of recommended vehicles. For a given trip or set of trips and a flexible list of candidate vehicles, the system outputs a recommendation based on a fast energy consumption prediction that accounts for cost, time and other fleet constraints. Examples of such constraints are operational costs, charging/refueling time delays, delivery window requirements, and truck payload needs.
}
\label{fig:summary}
\end{figure}

The rest of this article describes the development of a machine learning based vehicle-route assignment recommender system for MDHDT to efficiently identify cost effective technologies for different routes, cargo requirements, goods, energy cost, etc. from a total cost of ownership (TCO) perspective  (Figure \ref{fig:summary}) with very limited known internal road dynamics (i.e using high level route information).  With deployment goals in mind, the system needs to be lightweight, computationally efficient, accurate, and scalable for later integration into subsequent tools, allowing real-time querying. The objective of the paper is to provide and end-to-end deployable tool to support fleet decision making related to investments and technology usage.

\section{Recommender Systems}
\subsection{Background}
Recommender systems\textemdash algorithms that predict users' preferences among a large set of items based on their feedback\textemdash are widely used in many businesses today \cite{covington_deep_2016}, \cite{naumov_deep_2019}, \cite{gomez-uribe_netflix_2016}. A recommender system learns the systematic relationship between users and items based on past behavior as well as the item attributes involved. These systems are used to establish personalized systems that recommend items to the users likeliest to use or buy them. Table \ref{tab:PvsV} shows a direct similarity to our problem: We want to recommend vehicles for routes. Typically, recommender systems are built to deal with a large matrix of user\textendash item pairs, while our system operates similarly in a very large matrix of vehicle\textendash route possibilities. This approach also works well in a sparse setting, in which there is a large number of users and a big catalogue of items, but not all users interact with all items. In our case, the vehicle\textendash route design matrix is large, but not all vehicles use all routes. Our goal is to predict a rating, in our case TCO, to rate each vehicle\textendash route pairs.

\begin{table}[ht!]
\centering
\includegraphics[width=1\columnwidth]{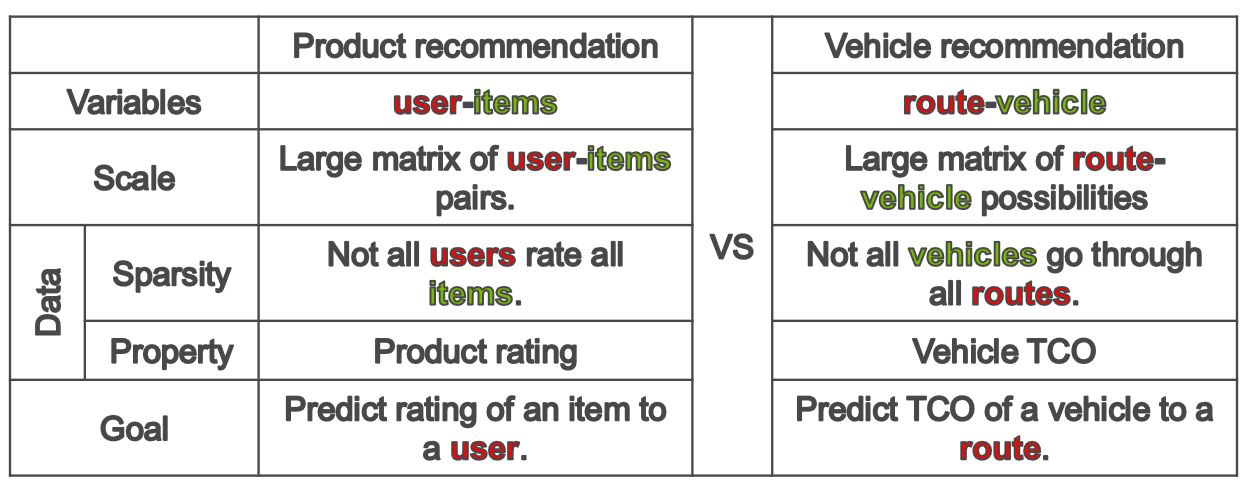}
\caption{Typical product recommendation compared to a vehicle recommendation}
\label{tab:PvsV}
\end{table}

Recommender systems can be built with variety of techniques and complexities. Collaborative filtering is one of the most popular and powerful traditional algorithms for a recommendation system. It uses a  sparse data matrix $R\in \mathbb{R}^{m\times n}$ in which each row represents a route and each column stands for a vehicle. A nonzero entry $r_{ui}$ of $R$ is the explicit feedback, i.e., rating, for the combination of route $u$ and vehicle $i$. Collaborative filtering leverages a method called low-rank matrix factorization, which computes the two low-rank matrices $V\in \mathbb{R}^{m \times k}$ and $P \in \mathbb{R}^{n\times k}$, $k<<n$ and $k<<m$, that contain the latent information on route\textendash vehicle interaction. The loss function of collaborative filtering to minimize is formulated as follows:
$$ \mathcal{L}= \sum_{(u,i)\in \Omega}(r_{u,i}-v_u^Tp_i)^2+\lambda ||V||_2+\lambda ||U||_2$$
where $v_u$ is the $u$-th row of $V$, $p_i$ is $i$-th column of $P^T$, $\Omega$ is the indices of nonzero entries of $R$, and $\lambda \geq 0$ is a regularizer parameter.

\subsection{Advances in Recommender Models and Related Work}
\label{sec:advRecSys}
 To generate more precise personalized recommendations, we can use  a deep neural network to input rich external features of both routes and vehicles, such as vehicle characteristics or road structures, into the algorithms.

Deep neural networks have been widely used in recommendation systems to learn the complex interactions of features from users and items. He et al. \cite{cf2} designed a neural-network version of collaborative filtering. They used an embedding layer to learn the latent factors $v_u$ and $p_i$, and they replaced the inner product with a more sophisticated fully connected neural network to learn the interaction of latent factors. Cheng et al. \cite{dnw} proposed a neural recommendation system they called the Wide \& Deep model that processes both raw and transformed features. It captures both linear and nonlinear interactions between features to learn and generalize users' preference for items by jointly training a wide single-layer neural network and a deep multi-layer neural network. With a raw input vector $x \in \mathcal{R}^d$, the model computes its predictions $\hat{y}$ as follows:
$$\hat{y}=\sigma\big(W_{\text{wide}}^T[x,\phi(x)] + W_{\text{deep}}^Ta^l + b \big)$$
where $\sigma(\cdot)$ is the sigmoid function, $a^l = f(W^la^{l-1}+b^l)$ is the output of $l$-th layer with $a^0$ being the compressed dense embedding of $x$, and $[x,\phi(x)]$ is the concatenation of $x$ and its cross-product transformation $\phi(x)$. \\

Based on the Wide \& Deep model, Guo et al. \cite{deepfm} proposed DeepFM, which replaces the wide model of Wide \& Deep with a more efficient factorization machine model. The factorization machine measures the interaction of feature $i$ and feature $j$ by the inner product of latent vectors $V_i$ and $V_j$ with a given dimension $k$. Hence, the output of the factorization machine is the following:
$$y_{\text{FM}}=W^Tx+\sum_{j_1=1}^{d}\sum_{j_1=j_2+1} \langle V_i,V_j \rangle x_{j_1}x_{j_2}$$
and DeepFM's outputs are the following:
$$\hat{y}=\sigma\big(y_{\text{FM}}+W_{\text{deep}}^Ta^l+b \big)$$
DeepFM improves the Wide \& Deep model as it directly takes raw input $x$ without the need for transformation. Wide \& Deep is essentially the ensemble of a linear model and a high-order nonlinear model, and DeepFM captures the additional order-2 interactions.
To learn users' preference for items, it is popular to integrate a model that learns linear and order-2 interactions of features with a deep high-order nonlinear model. Researchers have proposed several approaches to generating the order-2 interactions and having models learn them. Adding to the work of Guo et al., \cite{deepfm}, Lian et al. \cite{xdeepfm} introduced a convolutional neural network (CNN) approach called compressed interaction neural network (CIN) to learn these interactions, which they call extreme deep factorization machine (xDeepFM). xDeepFM encodes each feature in a vector of length $D$. Each raw datum is embedded as $X^0 \in \mathcal{R}^{m\times D}$, where \textit{m} is the number of features. It then computes the $k$ matrices of feature interactions through\textit{ k} layers and outputs $X^k\in \mathcal{R}^{H_k\times D}$. Each row of $X^k$ is computed recursively as follows:
$$X^k_h=\sum_{i=1}^{H_k}\sum_{j=1}^{m}X^{k-1}_i\odot X^0_j, \ \ 1\leq h \leq H_k$$
Based on each level of interaction, xDeepFM outputs the following:
$$p=[p^1,\cdots,p^k] \in \mathcal{R}^{\sum_{i=1}^kH_k}$$
where $p^l=X^l\mathbb{1}_l$ for $1\leq l \leq k $, and $\mathbb{1}$ is the vector of all \textit{l}s of length $H_l$. xDeepFM integrates CIN with a linear model and a deep neural network. It can be formulated as follows:
$$\hat{y}=\sigma\big(W^T_{\text{linear}}x+ W^T_{\text{CIN}}p+W^T_{\text{deep}}a^l+b\big)$$
Finally, here we introduce an additional approach to computing the interactions called \textquotedblleft cross layer" designed by Wang et al. \cite{cross1}. Given a input vector $x^0\in \mathcal{R}^d$, it computes its interaction with level $k$ recursively as shown:
$$x^k= x^0\odot (W^kx^{k-1}+b^k)+x^{k-1}$$ 
where $W^{k-1}\in \mathbb{R}^{d\times d}$ and $b^{k-1}\in \mathcal{R}^d$ are trainable weights and bias respectively. Here, we denote the cross layer operation as $\psi(\cdot)$, so the $k$-th order interactions generated by the cross layer are $x^k=\psi^{k}(x)$.

\section{Data and Learning Setup}

\begin{table}[ht!]
    \centering
    \begin{tabu}{c c}
        Symbol &  Description\\
        \tabucline[1pt]{-}
         $\mathcal{X}$& Sample space of data\\
         $\mathcal{X}_l$ & Collection of all trips with $l$ links\\
         $\mathcal{G}_c$ & \makecell{Collection of all trips \\ where vehicles have powertrain type $c$ }\\
         $\mathbf{X}_i$& Features of $i$-th trip (vehicle \& route)\\
         $\mathbf{y}_i$ &  2D consumption in $i$-th trip (fuel \& electric)\\
         $\mathbf{x}_t$ & Features of $t$-th link of a trip\\
         $\mathbf{u_t}$ & Route features of $t$-th link of a trip\\
         $\mathbf{v_t}$ & Vehicle features of $t$-th link of a trip\\
         $T_i$ & The number of links in $i$-th trip\\
         $\Tilde{T}$ & The number of links in an unspecified trip \\
         $f$& Vehicle model\\
         $\Theta$ & Overall weights of model\\
         $W$ & 2-dimensional weight of model \\
         $w$ & 1-dimensional weight of model\\
         $g \& \sigma$ & Activation functions\\
         $l(\cdot)$ & Loss function \\
         $\mathcal{L}(\Theta)$ & Objective function of training \\
         $[\mathbf{x}_1, \mathbf{x}_2]$ & Concatenation of vectors\\
         \tabucline[1pt]{-}
    \end{tabu}
    \caption{Table of mathematical notations in this work}
    \label{tab:notation}
\end{table}

\subsection{Data Generation Process}
\label{sec:DG}
\begin{figure}[ht!]
\centering
\includegraphics[width=\linewidth]{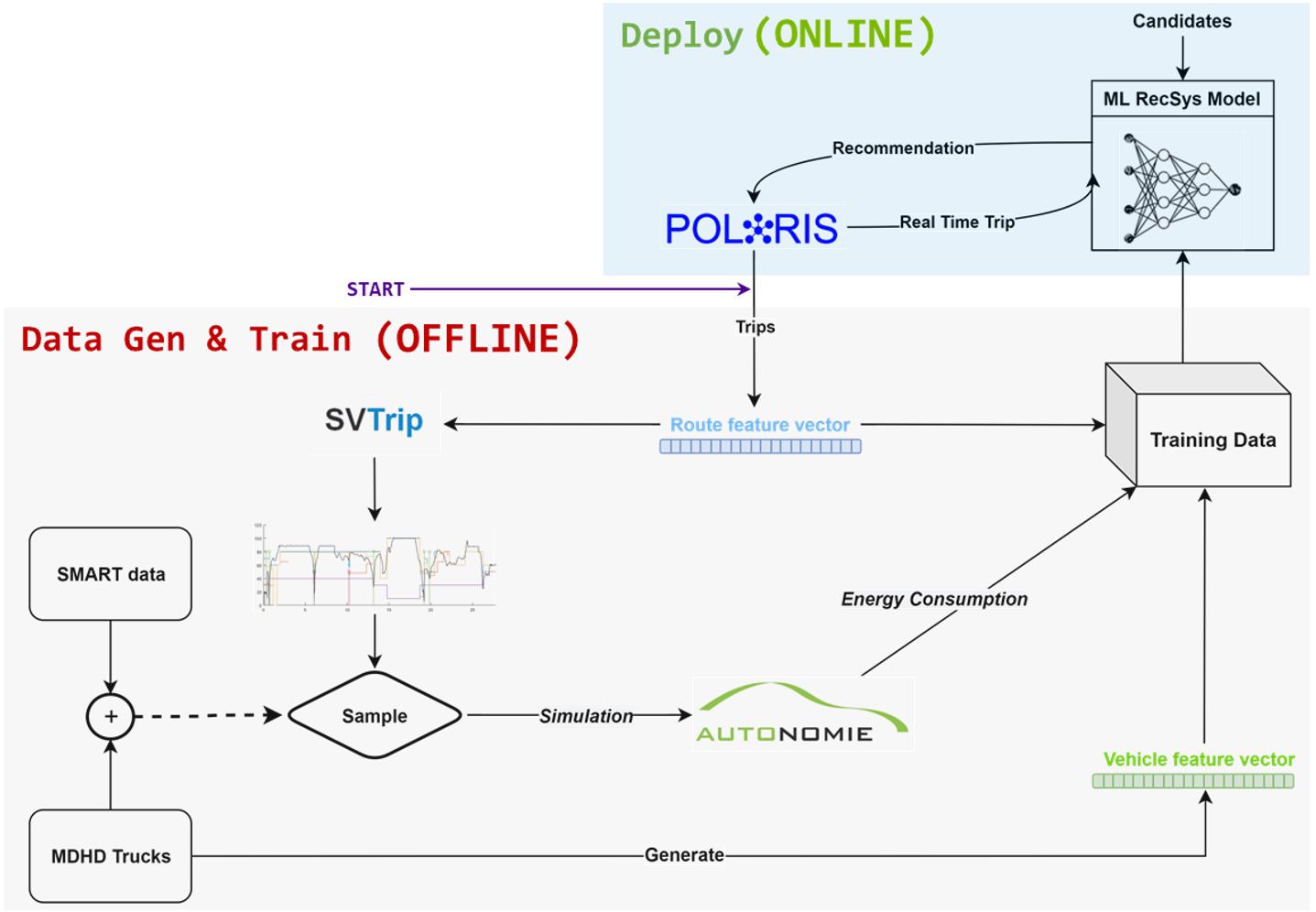}
\caption{Data generation and training process (grey box). After training, the model is deployed as a standalone model and integrated into the POLARIS Transportation System to make on-demand truck recommendations for new simulated fleets (light blue box).}
\label{fig:dgp}
\end{figure}

In order to determine the energy consumption of a vehicle in a large metropolitan area under varying conditions and scenarios, it is necessary to model the entire transportation system, including people and goods movement. Activity demand is generated to become the basis for the generation of trips. Trips generate traffic flow, which in turn affects average trip speed. Trip information can then be used along with powertrain information to inform energy consumption. Argonne National Laboratory has developed a complete workflow to study this complex and multi-dimensional system by combining several simulation tools \cite{freyermuth_energy_2019},  \cite{freyermuth_powertrain_2020}:  Autonomie,\footnote{Autonomie is a MATLAB-based software environment and framework for automotive modeling, control system design, simulation, and analysis.} SVTrip \cite{karbowski_trip_2014}, and POLARIS \cite{auld2016polaris}. The workflow is shown in Figure \ref{fig:dgp}.

POLARIS uses population and vehicle synthesis as well as activity demand generation and traffic flow. The freight transportation systems model is based on regional truck trip data \cite{noauthor_smart_nodate} with an e-commerce MDT module \cite{stinson2019}. The freight model is being updated to a fully agent-based freight model (named CRISTAL) with business firms and firm-specific delivery fleets \cite{STINSON2020771}, \cite{stinson_introducing_2021}. 

The route information is fed into SVTrip, which predicts the 1 Hz speed profile for each trip, which in turn is simulated with Autonomie to estimate the energy consumption of the transportation network for different vehicle technologies. This workflow operates offline as a data generator. A model is trained leveraging the data, deployed as a standalone model and integrated into the POLARIS Transportation System to make on-demand truck recommendations for new simulated fleets.

\subsection{Masking and Latent Energy Learning}
It is worth emphasizing that generated trips contain only basic information (expected average speed over link sequence) about the inner traffic dynamics. The goal is to learn energy consumption based on that incomplete summary-level dynamics information. Typically, high-resolution speed behavior is a key variable in estimating the energy profile, but in our setting the learning occurs without it: Our system learns the high-level route structure and vehicle\textendash feature interactions and how they correlate to the energy consumed. As a result, in the process shown in Figure \ref{fig:dgp}, the high-fidelity time series of vehicle speed dynamics is masked. The data used to train the energy model are vehicle parameters and high-level aggregated route information only. This approach means that future scenarios of this form could be constructed using readily available information such as that in HERE or Google Maps, this is consistent with information fleets have at their disposal in real world situations. POLARIS provides corresponding aggregated link level activities: Generated trips are on the level of detail of expected link average speed, link length, speed limits, and so on.
Although all internal dynamics that affect energy consumption are masked, we show that it is possible to learn aggregated-level energy consumption values quite accurately with a deep learning approach. When large-scale data is available, and with some tailored feature engineering, such a model is able to overcome latent information.

\subsection{The Data}
\label{sec:data}
The generated data contains over 3.5 million trips over 30,000 Chicago links. It accommodates a wide range of vehicle classes: class 3,  4,  6, and 8 trucks of different types, such as pickups\&delivery trucks, walk-ins, vans, boxes, long-hauls, etc. In addition, several powertrains were simulated with varying levels of electrification, component sizes, and technologies: from conventional to integrated starter generator (ISG) start-stop systems, hybrid electric vehicles (HEVs), plug-in hybrid electric vehicles (PHEVs), and pure battery electric vehicles (BEVs). Different fuel types (gasoline, diesel, etc.) and automation levels (no automation, partial automation, full automation) are also included.

\begin{figure}[ht!]
\includegraphics[width=\linewidth]{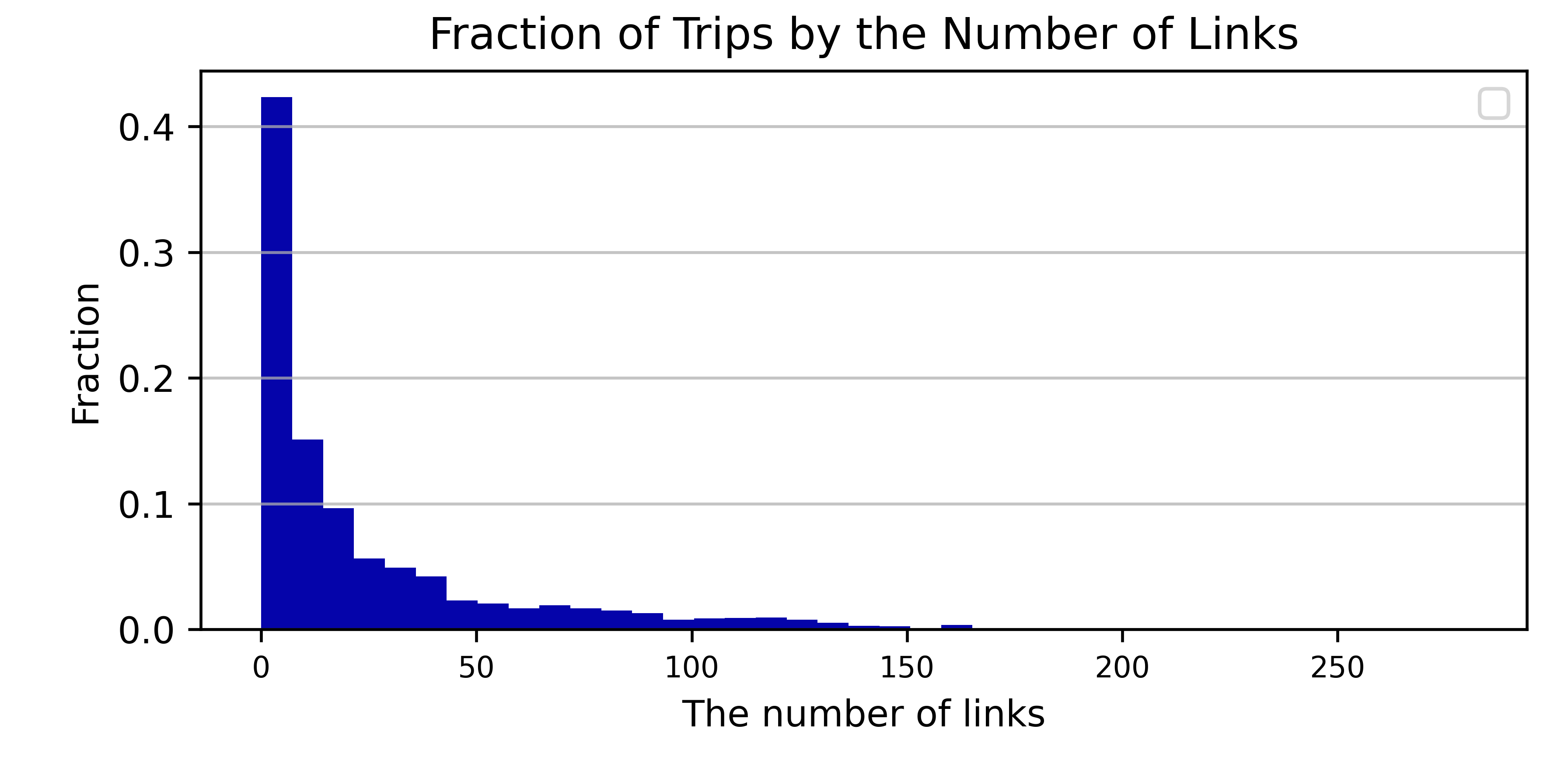}
\caption{Distribution of trips against the number of links in training data.}
\label{fig:distrbution_train}
\end{figure}

Let $\mathcal{X}= \{(\mathbf{X}_i,\mathbf{y}_i)\}_{i=1}^N$ and $\forall i$, $\mathbf{X}_i\in \mathcal{R}^{T_i\times D}$ be the dataset of trips with $T_i$ many links so that $\mathbf{X}_i=(\mathbf{x}_1,\cdots, \mathbf{x}_{T_i})$, where $\mathbf{x}_t=[\mathbf{u}_t,\mathbf{v}_t]\in \mathcal{R}^D$ is a link. We define $v_t\in \mathcal{R}^{D_1}$ and $u_t\in \mathcal{R}^{D_2}$ as the vehicle features and link features, respectively, with $D_1+D_2=D$ dimensional features. The labels $\mathbf{y}_i\in \mathcal{R}^{T_i\times 2}$ have two dimensions: the electricity consumption and fuel consumption of vehicles on each link. The entire sample space can be partitioned in many ways based on classes of a certain feature. We are mostly interested in two  partitions: powertrain type and number of links. Because the number of links in a trip is an arbitrary positive integer, by the number of links in a trip, we can naturally partition our sample space into countably many disjoint subspaces:
$$\mathcal{X}=\bigsqcup_{l\geq 0}\mathcal{X}_l, \  \ \mathcal{X}_l=\mathcal{R}^{l\times D}\times \mathcal{R}^{l\times 2}$$ $\mathcal{X}_l$ is the collection of all possible trips that contains exactly $l$ links. Then again, the sample space can be partitioned by the powertrain type:
$$\mathcal{X}=\bigsqcup_{c=0}^{4}\mathcal{G}_c$$
where the subscript stands for one of the five powertrain types, i.e., 
$\{0: \text{BEV}, 1: \text{Conv}, 2: \text{ISG}, 3: \text{HEV}, 4: \text{PHEV} \}$. In this work, we will focus on how our model performs on each powertrain, as our goal is to make powertrain selection recommendations for given trips. As shown in Figure \ref{fig:distrbution_train}, the distribution of trips against  the number of links is heavily skewed to the right. Most trips have fewer than 50 links. Table \ref{tab:Train_stats} shows key distribution statistics over the energy consumption of diverse powertrains. Fuel consumption mass values are converted to gasoline equivalent values in grams for a uniform comparison of the different fuel types, while electric consumption is in Wh. Note that the mean is significantly higher than the median, which indicates a large variance in consumption values. Therefore, we will analyze the errors against the number of links in trips to see whether the model is biased toward short trips.

\begin{table}[]
\centering
\begin{tabular}{crrrrrr}
                                       & \multicolumn{6}{c}{\textbf{Fuel Consumed {(}g{)}}}                                                                                   \\ \hline
\multicolumn{1}{c|}{\textbf{Veh Type}} & \textbf{count}       & \textbf{mean}        & \textbf{std}         & \textbf{min}         & \textbf{50\%}        & \textbf{max}         \\ \hline
\multicolumn{1}{c|}{\textbf{Conv}}     & 686656               & 7185.0               & 10917.4              & 7.0                  & 2036.6               & 108668.4             \\
\multicolumn{1}{c|}{\textbf{HEV}}      & 513508               & 6871.1               & 10542.4              & 0.0                  & 1858.0               & 103570.8             \\
\multicolumn{1}{c|}{\textbf{ISG}}      & 514992               & 6939.3               & 10641.8              & 3.2                  & 1904.9               & 104763.5             \\
\multicolumn{1}{c|}{\textbf{PHEV}}     & 514746               & 2225.3               & 5124.8               & 0.0                  &   41.6                 &  49139.2              \\
                                       & \multicolumn{1}{c}{} & \multicolumn{1}{r}{} & \multicolumn{1}{r}{} & \multicolumn{1}{r}{} & \multicolumn{1}{r}{} & \multicolumn{1}{r}{} \\
                                       & \multicolumn{6}{c}{\textbf{Electric Consumption {(}Wh{)}}}                                                                            \\ \hline
\multicolumn{1}{c|}{\textbf{Veh Type}} & \textbf{count}       & \textbf{mean}        & \textbf{std}         & \textbf{min}         & \textbf{50\%}        & \textbf{max}         \\ \hline
\multicolumn{1}{c|}{\textbf{BEV}}      & 384530               & 36570.2              & 58431.4              & 13.8                 & 9280.6               & 599554.5             \\
\multicolumn{1}{c|}{\textbf{PHEV}}     & 514746               & 25845.8              & 36995.5              & 16.8                 & 7912.0               & 374345.6             \\ 
\end{tabular}
\caption{Statistics of the training set at the trip level}
\label{tab:Train_stats}
\end{table}

In our dataset, we have $D_1=20$ the dimensionality of the vehicle features and $D_2=60$ the dimensionality of the link features, so that $D=80$. $D_1$ includes vehicle data such as vehicle class, weight, battery size, frontal area, engine power, etc. Among the $D_2=60$ only 7 are direct use of raw route features from macroscopic level data, the rest are carefully engineered/derived from them. Examples of raw data are: link ID, the link entering time of the vehicle, the link length, the expected duration of stop on link, the expected duration of travel on link, the expected average speed on link and the speed limit on link. Those are direct output of POLARIS. Examples of manually engineered features are: delta link average speed of surrounding links, calculated proxies of congestion by comparing speeds to speed limits, length of previous and subsequent links, difference in speed between links over travel time as proxy to global acceleration, etc. Such calculated macroscopic level dynamics are aftereffect of more granular microscopic changes. Designed features that capture link sequence dependencies will explicitly guide and inform the model training to learn microscopic latent behavior. We note that the vehicle features used in the model are limited to major vehicle attributes that are in general considered available and known, or can be collected as shown in \cite{XAI_moawad}.

Trips have different lengths, and zero padding is a common way to equalize trip data points to the same shape. However, in our case, the trip length varies from 1 to 283, and padding jeopardizes the accuracy of the model. To better train our models, we experimented with different ways of batching the link sequence data, which varied  by trip length. First, we grouped the data by similar trip length (with shuffling at the end of each epoch to introduce noise) to minimize padding within a batch. After grouping by similar trip length, we further experimented with quantile bucketing and binning of different sizes. That is, if $T=\max_{1\leq i \leq N}T_i$, we discretize the range of trip length into levels $0= L_0<L_1<\cdots < L_m = T$. Based on this discretization, we partitioned the dataset into disjoint groups $\mathcal{X}=\bigsqcup_{k=1}^{m}\mathcal{G}_k$, where $X_i\in \mathcal{G}_k$ if $T_i\in (L_{k-1},L_{k}]$. Then, for each group $\mathcal{G}_k$, we mapped all data points in $\mathcal{G}_k$ into $\mathcal{R}^{L_k\times D}$ by padding with zeros.

Next, at the expense of less noise during training, we also grouped the sequence data by \textit{exactly} the same trip length to avoid padding completely. This approach yielded the best result. 
 
As explained earlier, if we treat each vehicle as an item and each link as a user, we turn our recommendation problem into a classic item\textendash user recommendation. We recommend vehicles based on a trip, which is a sequence of links. In this setting, we want to recommend a vehicle to maximize the expected feedback given a ordered sequence of links. Our task is more challenging than evaluating each item\textendash user pair. We need to model beyond the similarity of users and items, as the energy of each link will have an impact on the following link.

\section{Our Model}
\subsection{Setup}
The key to building a recommender system for our task is being able to accurately estimate the energy consumption of a large vehicle given varying roads, trips, and driving styles. Although we have features that seem to directly affect energy consumption, such as the trip length, average speed, vehicle weight, etc., there are significant difficulties in modeling the energy consumption explicitly because it is hard to capture the interactions between links. For example, since the fuel consumption for acceleration and slowing down are very different, even with the same \textit{average} speed a vehicle's energy consumption on a link varies depending on how it enters and leaves the link. A neural network can learn such latent information from a sufficiently large and rich dataset. Inspired by the neural recommender systems mentioned above, we designed a novel way to learn a vehicle's expected energy consumption on a given trip: We use the idea of an ensemble model and integrate models with their distinct functions to accurately predict energy consumption.

\subsection{Structure}
As mentioned in the previous section, the energy consumption of a vehicle on a link depends on the preceding and following links. For that  portion of the model, we leverage a recurrent neural network with long short term memory (LSTM) cells \cite{hochreiter1997long} of hidden size 128, stacked with a fully connected neural network to learn the sequential dependencies of link consumption. Given a input trip $\mathbf{X}$ with length $\Tilde{T}$, the output is formulated as follows:
\begin{equation}
\begin{aligned}
&\mathbf{H} = LSTM(\mathbf{X})\\
&\Tilde{y}_{rnn}=g(\mathbf{H}w_{rnn}+b_{rnn})
\end{aligned}  
\end{equation}
where $\mathbf{H}\in \mathcal{R}^{\Tilde{T}\times 128}$, $w_{rnn}\in \mathcal{R}^{128}$, and the output $\Tilde{y}_{rnn}\in R^{\Tilde{T}}$. $g(\cdot)$ is the rectified linear activation function. 

While the LSTM captures the interaction between links, we built two models that learn how the characteristics of the vehicle and route itself affect the energy consumption. First, we have a linear model that learns from direct features  and the order-2 interaction of features. There are known feature relationships that directly and unambiguously correlate with energy (e.g., link length), which we want to explicitly capture to inform the model during training. To compute order-2 interactions, we use the cross layer from \cite{cross1} mentioned in Section \ref{sec:advRecSys}. Its advantage is its high efficiency in computation and memory. If we directly generate a flat vector of order-2 interactions of all features for a link, this brute-force approach will give a vector of length ${D \choose 2} + D$ (cross terms plus squared terms), which makes computation very expensive. The cross layer operator outputs a vector of length $D$ that contains all the cross terms and squared terms with trainable weights. It allows our model to capture the order-2 interactions with low cost. For each trip, the cross layer operator is applied link-wise, i.e., $\psi(\mathbf{X})=\big(\psi(\mathbf{x}_1),\cdots,\psi(\mathbf{x}_{\Tilde{T}})\big)$. Our linear model is formulated as follows:
\begin{equation}
    \Tilde{y}_{linear}=g(\mathbf{X}w_{l_1}+\psi(\mathbf{X})w_{l_2}+b_l)
\end{equation}
where $w_{l_1},w_{l_2},b_l\in \mathcal{R}^{D}$.

Finally, we integrate a personalization component to learn the preference for each vehicle on each link. We have a deep model that resembles  neural collaborative filtering. Based on \cite{cf2}, we used two embedding layers to map vehicle ID and link ID into two latent vectors, $v_u^t$ and $p_i^t$, of dimension 32. Then, we concatenate the latent vectors and pass the result to a multi-layer network to train. For this deep model, we have the following:
$$\mathbf{a}^0=(\mathbf{a}_1^0,\cdots,\mathbf{a}_{\Tilde{T}}^0)$$
with link-wise concatenated embedding $\mathbf{a}_t^0=[v_u^t, p_i^t]\in \mathcal{R}^{64}$ for $1\leq t \leq \Tilde{T}$. For each layer, we have the following:
$$\mathbf{a}^j = f(W_j\mathbf{a}^{j-1}+b_j)=(f(W_j\mathbf{a}_j^{j-1}+b_j),\cdots,f(W_j\mathbf{a}_{\Tilde{T}}^{j-1}+b_j))$$
where $W_j \in \mathcal{R}^{H_j\times H_{j-1}}$ with $H_0=32$ and $H_l=1$. As we have one dense layer ($W_l\in \mathcal{R}^{1\times H_{l-1}}$) in the end, we may define $w_l=W_l^T \in \mathcal{R}^{H_{l-1}}$ and formulate the output $\mathbf{a}^l$ as follows:
\begin{equation}
    y_{deep} = f(\mathbf{a}^{l-1}w_l+b_l)
\end{equation}
We integrate three models to predict the link-level consumption:
\begin{equation}
    \hat{y} = \sigma\big(y_{rnn}+y_{linear}+y_{deep} \big)
\end{equation}

The key to this modeling is that, because vehicles don't enter or exit links the same way, similar (or even identical) links don't necessarily yield similar energy profiles. In a given trip, a route is composed of a sequence of links, and the energy consumed depends on previous and subsequent links structures and characteristics. Modeling the entire sequence informs link-level latent internal dynamics, provided a large dataset of trips (i.e. sequences) is available. In other words, such model is able to retrieve microscopic latent behavior.

\subsection{Loss Function and Training}
Our final goal is to recommend the vehicle with the lowest potential energy consumption given a new trip $\mathbf{X}$. Our recommendation will be based on the estimated consumption of the entire trip, $\hat{y}^{i}_{trip}=\sum_{t=1}^{\Tilde{T}}\hat{y}_{t,i}$. This new trip can be a random combination of existing (adjacent) links and unknown to our model. Therefore, we require our model to be accurate in both link-level and trip-level consumption, so the model can learn both interactions of vehicles and routes and the impact of the sequential effect of links. To further improve the generalization ability of our model, we impose an additional cumulative link loss that requires the model to be precise on partial trips up to each link. The loss is formulated as follows:  
\begin{equation}
\label{eqn:loss}
\begin{aligned}
    \mathcal{L}(\Theta)=  \frac{1}{N}\sum_{i=1}^{N}\Bigg( & \frac{1}{T_i}\sum_{l=1}^{T_i}\Big[\sum_{t=1}^{l}\big(y_{t,i}-f(\Theta, \mathbf{x}_{t,i})\big)\Big]^2\\
    &+\Big[\sum_{t=1}^{T_i}\big(y_{t,i}-f(\Theta,\mathbf{x}_{t,i})\big)\Big]^2 \\
    &+\frac{1}{T_i}\sum_{t=1}^{T_i}\big(y_{t,i}-f(\Theta,\mathbf{x}_{t,i})\big)^2 \Bigg)
\end{aligned}
\end{equation}
We search for the optimal solution: $$\Theta^* = \argmin_{\Theta} \mathcal{L}(f(\Theta)
)$$ through back-propagation gradient descent.  $C_l=\frac{1}{T_i}\big[\sum_{t=1}^{l}\big(y_{t,i}-f(\Theta, \mathbf{x}_{t,i})\big)\big]^2$ evaluates the model's performance on the partial trip up to the $l$-th link. We sum $C_l$ over all $l$ to find the cumulative link loss, $C_{sum}=\sum_{l=1}^{T_i}C_l$, which evaluates the model's predictions on partial trips up to each link. The trip level loss $C_{trip}=\big[\sum_{t=1}^{T_i}\big(y_{t,i}-f(\Theta,\mathbf{x}_{t,i})\big)\big]^2=(\hat{y}^{i}_{trip}-y^{i}_{trip})^2$ evaluates the model's prediction of the final consumption for the entire trip, and $C_{link}=\frac{1}{T_i}\sum_{t=1}^{T_i}\big(y_{t,i}-f(\Theta,\mathbf{x}_{t,i})$ measures the model's performance on each link. As we mentioned in the previous section, we group trips based on their length without padding. Trips in the same batch share a common $T_i$, so objective function \eqref{eqn:loss} is precise in training.  \\

For training and tuning purposes, we leveraged multiple graphics processing units (GPUs) spread across multiple machines via MultiWorkerMirroredStrategy, which implements synchronous distributed training across multiple workers, each with potentially multiple GPUs. With the help of roughly 30 machines, each with at least one GPU, we were able to decrease the model training and tuning time for a dataset of this scale. Our hyperparameter search space contained different hyperparameters spanning different areas of model development, including the different pre-processing approaches discussed in Section \ref{sec:data}, model selection and design, but also training strategies to include/exclude or freeze the weights of the different sub-component models in the ensemble at different times during the training phase. In the remaining sections, we present the results of the best trained model.

\section{Results and Analysis}
\subsection{Metrics}
We introduce several metrics to evaluate model performance. The most direct measurement for a regression problem is the mean absolute error (MAE), formulated as follows: $$\text{MAE}=\frac{1}{N}\sum_{i=1}^{N}\big|y^{i}_{\text{trip}}- \hat{y}^{i}_{\text{trip}}\big|$$
And Root mean square error (RMSE): 
$$\text{RMSE}= \sqrt{\frac{1}{N}\sum_{i=1}^{N}\big(y^{i}_{\text{trip}}- \hat{y}^{i}_{\text{trip}}\big)^2}$$
Both MAE and RMSE average the size of errors, while RMSE is sensitive to the variance of errors. Let $e_i=|y^{i}_{\text{trip}}- \hat{y}^{i}_{\text{trip}}|$ and $X$ be the discrete uniform random variable on set $\{e_i\}_{i=1}^{N}$. One can easily show that 
$$\text{RMSE}^2=\mathbb{E}[X^2]=Var(X)+(\mathbb{E}[X])^2=Var(X)+\text{MAE}^2$$
That is, the difference between RMSE and MAE measures the variance of the errors. We use RMSE and MAE to evaluate the performance of our models in terms of both size and variance of errors. \\
On the other hand, RMSE and MAE have their flaws. There are some long trips in our dataset where energy consumption can reach tens of thousands of grams or kWh. In that case, a  1\% error in prediction will lead to a large RMSE and MAE. Therefore, we also evaluate our model with two additional metrics: mean absolute percentage error (MAPE) and its variant mean arc-tangent absolute percentage error (MAAPE).\\
MAPE is defined as shown: 
$$\text{MAPE}= \frac{1}{N}\sum_{i=1}^{N}\Big|\frac{y^{i}_{\text{trip}}- \hat{y}^{i}_{\text{trip}}}{y^{i}_{\text{trip}}}\Big|$$
which is the mean of absolute percentage error (APE), i.e., $\Big|\frac{y^{i}_{\text{trip}}- \hat{y}^{i}_{\text{trip}}}{y^{i}_{\text{trip}}}\Big|$. In other words, it expresses the error of the prediction in terms of a percentage. MAPE's  drawback is that it is extremely sensitive to outliers, especially when 
$y^{i}_{\text{trip}}$ is small. One outlier can make MAPE very large, and it is impossible to tell the model's performance on the entire dataset from this large number. In order to fix this issue, Kim and Kim \cite{maape} introduced MAAPE, formulated as follows:
$$\text{MAAPE}= \frac{1}{N}\sum_{i=1}^{N}\arctan \Big|\frac{y^{i}_{\text{trip}}- \hat{y}^{i}_{\text{trip}}}{y^{i}_{\text{trip}}}\Big|$$
MAAPE averages the arc-tangent absolute error (AAPE), $\arctan \Big|\frac{y^{i}_{\text{trip}}- \hat{y}^{i}_{\text{trip}}}{y^{i}_{\text{trip}}}\Big|$.  Each absolute percentage error is bounded by $\frac{\pi}{2}$ in MAAPE, so the value of MAAPE will not be significantly affected by a small number of outliers when the sample size is large. AAPE also approximates absolute percentage error well for small errors. From the Taylor series of $\arctan{x}$, we see that $|\arctan{x}-x|=\mathcal{O}(\epsilon^3)$ for $x \in [0,\epsilon]$. As a result, MAAPE can give us a better idea of the overall performance of the model when there are outliers in errors.

\subsection{Validation Data}
\begin{table}[]
\centering
\resizebox{\linewidth}{!}{%
\begin{tabular}{crrrrrrr}
                                       & \multicolumn{7}{c}{\textbf{Fuel Consumed {(}g{)}}}                                                                                                                                                                                            \\ \hline
\multicolumn{1}{c|}{\textbf{Veh Type}} & \textbf{mean}               & \multicolumn{1}{c}{\textbf{std}} & \multicolumn{1}{c}{\textbf{min}} & \multicolumn{1}{c}{\textbf{25\%}} & \multicolumn{1}{c}{\textbf{50\%}} & \multicolumn{1}{c}{\textbf{75\%}} & \multicolumn{1}{c}{\textbf{max}} \\ \hline
\multicolumn{1}{c|}{\textbf{Conv}}     & 7647.6                      & \multicolumn{1}{c}{11645.8}      & \multicolumn{1}{c}{18.7}         & \multicolumn{1}{c}{579.7}         & \multicolumn{1}{c}{2044.5}        & \multicolumn{1}{c}{9828.8}        & \multicolumn{1}{c}{83649.5}      \\
\multicolumn{1}{c|}{\textbf{HEV}}      & \multicolumn{1}{l}{6956.4}  & 10679.7                          & 0                                & 525.1                             & 1867.5                            & 8797.6                            & 77315.8                          \\
\multicolumn{1}{c|}{\textbf{ISG}}      & \multicolumn{1}{l}{7133.4}  & 10927.8                          & 11.5                             & 540.8                             & 1903.6                            & 9167.6                            & 93428.2                          \\
\multicolumn{1}{c|}{\textbf{PHEV}}     & \multicolumn{1}{l}{2360.8}  & 5324                             & 0                                & 0                                 & 53.6                              & 1144.6                            & 43939.4                          \\
                                       &                             & \multicolumn{1}{c}{}             & \multicolumn{1}{c}{}             & \multicolumn{1}{c}{}              & \multicolumn{1}{c}{}              & \multicolumn{1}{c}{}              & \multicolumn{1}{c}{}             \\
                                       & \multicolumn{7}{c}{\textbf{Electric Consumption {(}Wh{)}}}                                                                                                                                                                                     \\ \hline
\multicolumn{1}{c|}{\textbf{Veh Type}} & \textbf{mean}               & \multicolumn{1}{c}{\textbf{std}} & \multicolumn{1}{c}{\textbf{min}} & \multicolumn{1}{c}{\textbf{25\%}} & \multicolumn{1}{c}{\textbf{50\%}} & \multicolumn{1}{c}{\textbf{75\%}} & \multicolumn{1}{c}{\textbf{max}} \\ \hline
\multicolumn{1}{c|}{\textbf{BEV}}      & \multicolumn{1}{l}{36655.9} & 59668.5                          & 39.7                             & 2449.6                            & 8734.4                            & 43080.2                           & 449268.1                         \\
\multicolumn{1}{c|}{\textbf{PHEV}}     & \multicolumn{1}{l}{26566.9} & 38497.6                          & 53.4                             & 2203.1                            & 7702.9                            & 39004.4                           & 293764.2                        
\end{tabular}%
}
\caption{Statistics of the validation set}
\label{tab:Val_stats}
\end{table}

The validation set was generated by a random 20\% split of the entire data. An import feature we focus on is the number of links in each trip. Since the model predicts the consumption on each link and outputs the summation as the prediction for a trip, the number of links in a trip can potentially play a role in the model's accuracy. As shown in Figure \ref{fig:distrbution_train}, the distribution is heavily skewed to the right; most trips have fewer than 50 links, and there are few trips with more than 100 links. 

The large number of links does not necessarily mean that the trip is long, as the length of a link varies across a wide range. However, we find that the number of links is strongly positively correlated to the trip length. The correlation coefficient of the two features is 0.83, which means a trip with high number of links is very likely to be long. We identified only a few trips that are exceptionally long and have high energy consumption.

\subsection{Residual Analysis and Outliers}
We first measured error by size, as shown in Table \ref{tab:magnitude}. As the validation data can be partitioned based on powertrain type, we also computed the error on each vehicle type, as shown. The model has fairly low MAE and RMSE on both fuel and electric consumption when compared to the range of the true values from the validation set statistics shown in Table \ref{tab:Val_stats}.

\begin{table}[]
\centering
\begin{tabular}{crr}
                                       & \multicolumn{2}{c}{\textbf{Fuel Consumed {(}g{)}}}        \\ \hline
\multicolumn{1}{c|}{\textbf{Veh Type}} & \textbf{MAE}                  & \textbf{RMSE}                \\ \hline
\multicolumn{1}{c|}{\textbf{Conv}}     & 24.8                          & 50.9                         \\
\multicolumn{1}{c|}{\textbf{HEV}}      & 34.8                          & 87.3                         \\
\multicolumn{1}{c|}{\textbf{ISG}}      & 188.2                         & 1582.2                       \\
\multicolumn{1}{c|}{\textbf{PHEV}}     & 97.1                          & 565.1                        \\
                                       &                               &                              \\
                                       & \multicolumn{2}{c}{\textbf{Electric Consumption {(}Wh{)}}} \\ \hline
\multicolumn{1}{c|}{\textbf{Veh Type}} & \textbf{MAE}                  & \textbf{RMSE}                \\ \hline
\multicolumn{1}{c|}{\textbf{BEV}}      & 110.4                         & 255.9                        \\
\multicolumn{1}{c|}{\textbf{PHEV}}     & 254.4                         & 655.7                        \\ \hline
\end{tabular}
    \caption{Mean absolute error (MAE) and root mean squared error (RMSE) of trip energy consumption }
    \label{tab:magnitude}
\end{table}

\begin{figure*}[ht!]
\includegraphics[width=\textwidth]{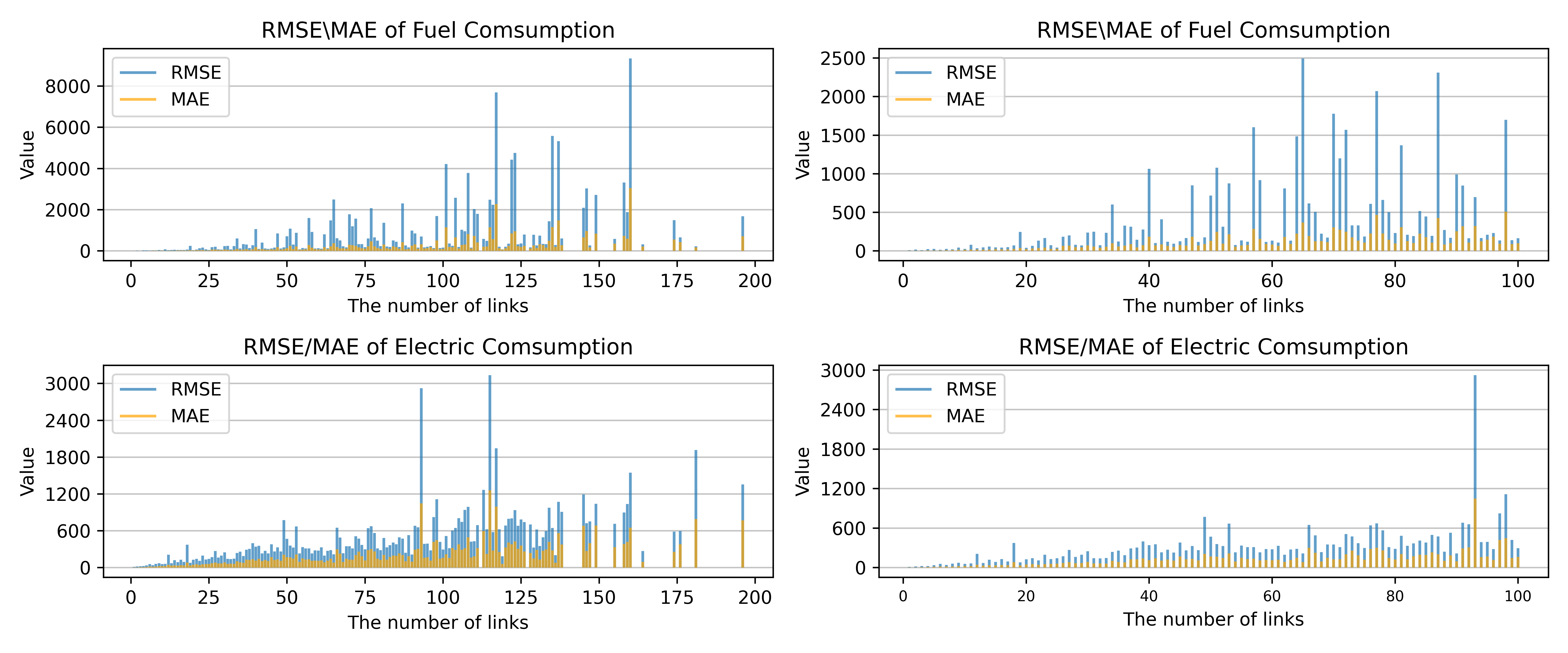}
\caption{MAE and RMSE of trips by the number of links in the trip. The left-hand section shows plots of all trips of varying number of links. The right-hand side section shows trips with fewer than 100 links.}
\label{fig:RMSE_and_MAE}
\end{figure*}

\begin{figure}[ht!]
\includegraphics[width=\linewidth]{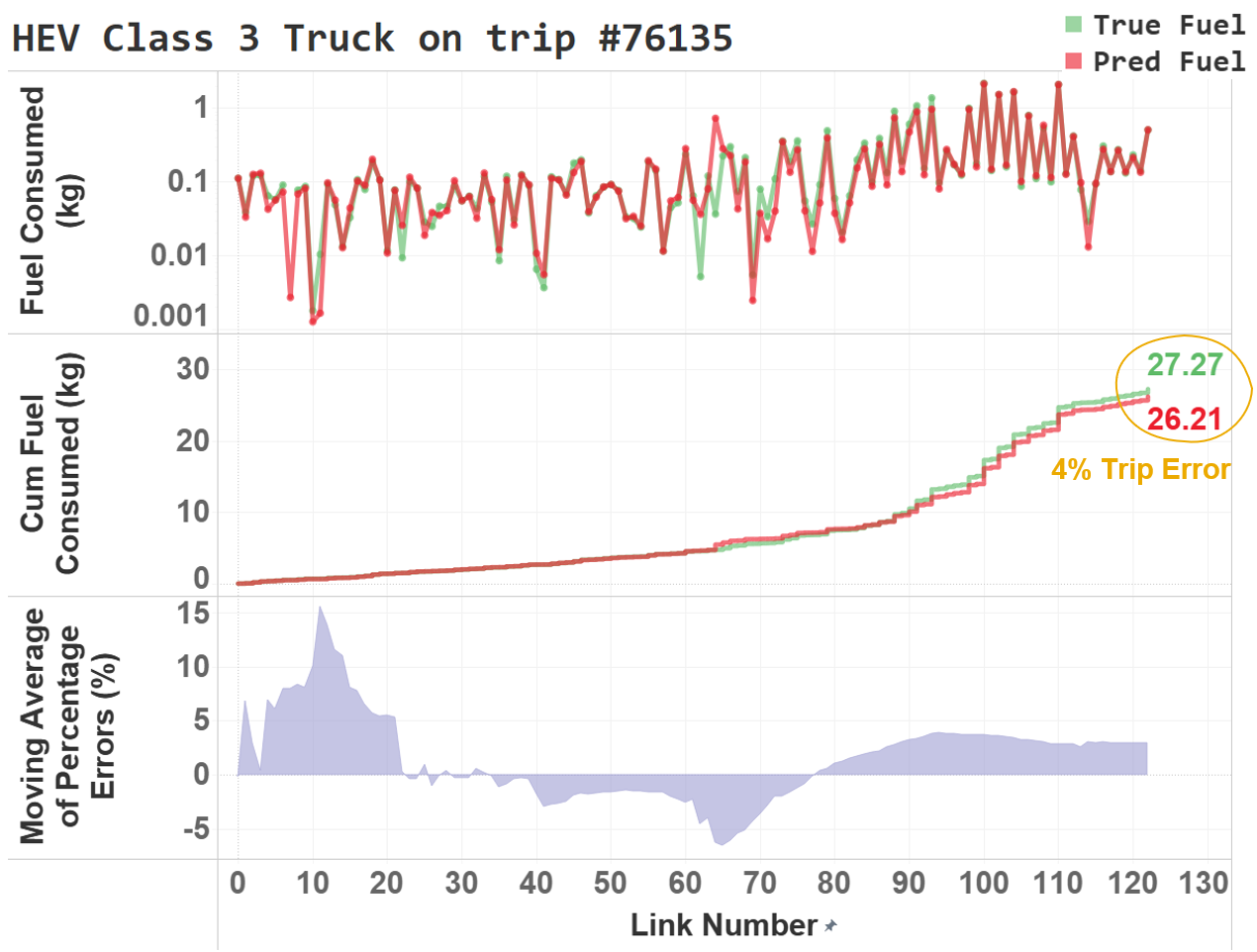}
\caption{Link level prediction and cumulative prediction of a trip.}
\label{fig:a_trip}
\end{figure}

The large difference between MAE and RMSE estimates indicates that the variance of the errors is large. This behavior suggests that the distribution of errors is skewed and extreme outliers might appear. Since the dataset is seriously imbalanced in terms of the number of links as well as the energy consumption values, we investigated the error size by the number of links. The RMSE and MAE over all trips with $l$ many links are computed as follows:
$$\text{RMSE}_l=\sqrt{\frac{1}{|\mathcal{X}_l|}\sum_{(\mathbf{X}_i,\mathbf{y}_i)\in\mathcal{X}_l}\big(f(\mathbf{X}_i)-y^{i}_{trip}\big)^2}$$
$$\text{MAE}_l=\frac{1}{|\mathcal{X}_l|}\sum_{(\mathbf{X}_i,\mathbf{y}_i)\in\mathcal{X}_l}\big|f(\mathbf{X}_i)-y^{i}_{trip}\big|$$
RMSE and MAE by number of links are shown in Figure \ref{fig:RMSE_and_MAE}. It is clear that $\text{RMSE}_l$ and $\text{MAE}_l$ increase as $l$ increases. However, it does not necessarily mean that the model performs worse on long trips. For example, Figure \ref{fig:a_trip} shows the model's prediction for a Class 3 hybrid truck's trip. This trip has 121 links and is over 170 kilometers long. We see that the model predicts the consumption on each link quite precisely. The truck consumes 27.27 kg of fuel on the entire trip, while the model predicts the consumption to be 26.21. The model underestimates the consumption by only 4\%, but this trip contributes an error of over 1 kg (1000 g) to $\text{MAE}$ and $\text{MAE}_{121}$, which can appear to be a large number in absolute.

\begin{figure*}[ht!]
\includegraphics[width=\textwidth]{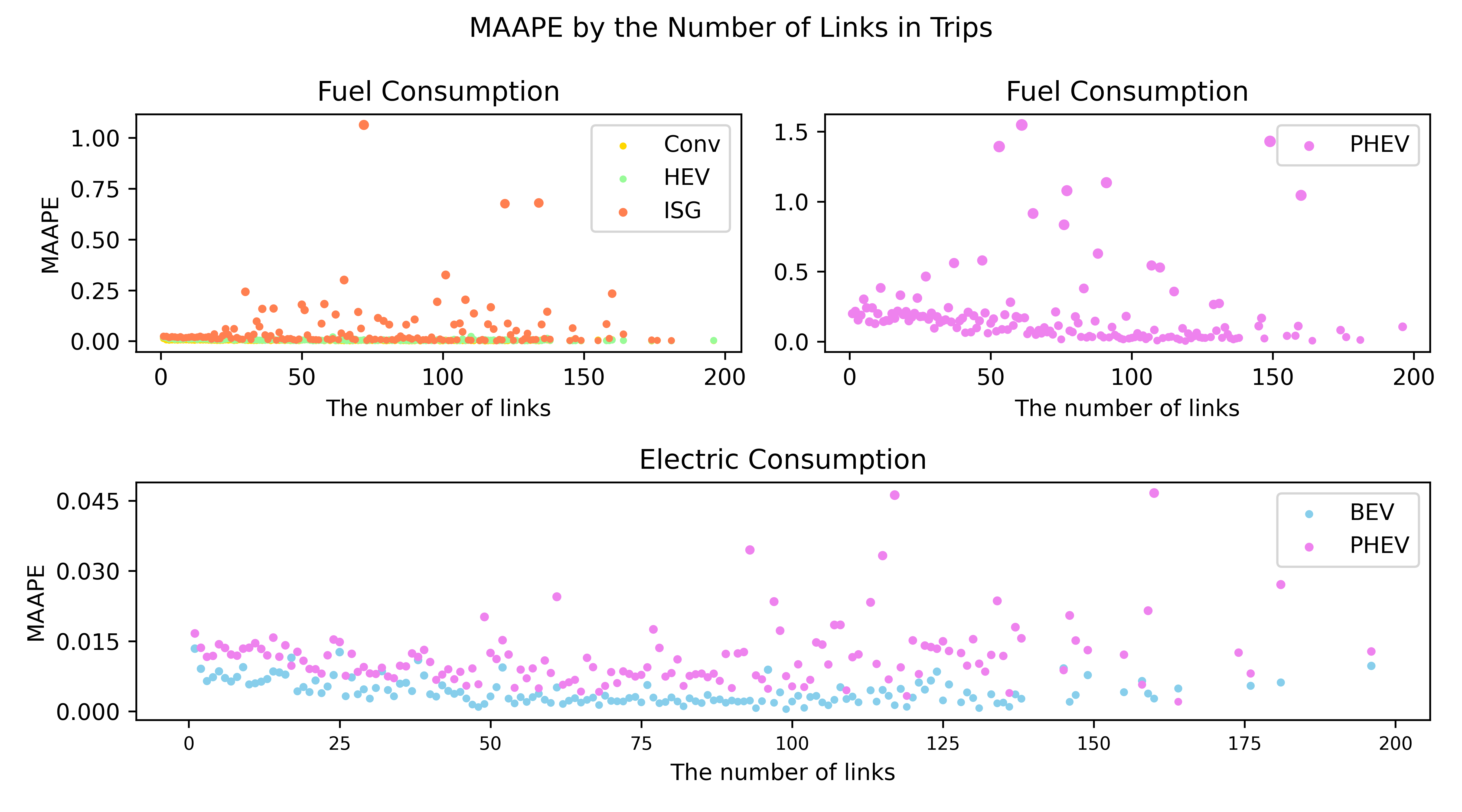}
\caption{MAAPE of each powertrain by the number of links.}
\label{fig:MAAPE}
\end{figure*}

To further investigate the model's performance on trips with different numbers of links, we computed the $\text{MAPE}_l$ and $\text{MAAPE}_l$. However, the MAPE and MAAPE are not defined for sets in which the corresponding consumption is zero (e.g., fuel consumption for a BEV is zero). Therefore, we computed $\text{MAAPE}_l$ for each powertrain type separately. Defining $\mathcal{G}^+_c=\{(\mathbf{X}_i,\mathbf{y}_i)\in\mathcal{G}_c|y^{i}_{trip}\not = 0\}$, we compute the following:

$$\text{MAAPE}^c_l=\frac{1}{|\mathcal{X}_l\cap \mathcal{G}^+_c|}\sum_{(\mathbf{X}_i,\mathbf{y}_i)\in \mathcal{X}_l\cap \mathcal{G}^+_c}\arctan\Big|\frac{f(\mathbf{X_i})-y^{i}_{trip}}{y^{i}_{trip}}\Big|$$

We have $c\in \{1,2,3,4\}$ for fuel consumption and $c\in \{0,4\}$ for electric consumption. We present the result in Figure \ref{fig:MAAPE} in a plot of MAAPE against the number of links for each powertrain. For electric consumption, we note that the BEV is not sensitive to the number of links, while $\text{MAAPE}^4_l$ (PHEV) increases slightly as $l$ increases. For fuel consumption, the errors of conventionals and HEVs are not very sensitive to the number of links either. $\text{MAAPE}^3_l$ increases as the $l$ increases.

\begin{table}[ht!]
\centering

    \begin{tabular}{c c c c c c} 
        \hline
        Consumption  &25th & 50th & 75th & 90th & 95th\\ 
        \hline 
        Fuel   & 0.003 & 0.006 & 0.017 & 0.057 & 0.157\\
        Electric   & 0.002 & 0.004 & 0.008 & 0.015 & 0.022 \\ 
        \hline 
    \end{tabular}
\caption{AAPE percentiles of fuel consumption by non-electric vehicles and APE percentiles of electric consumption by BEVs and PHEVs}
\label{tab:percentiles}
\end{table}
We computed the AAPE of fuel consumption over all trips in $\bigsqcup_{c=1}^4 \mathcal{G}^+_c$  and the AAPE of electricity consumption over all trips in $\mathcal{G}^+_{0}\bigsqcup\mathcal{G}^+_{4}$. We list the percentiles of those errors in Table \ref{tab:percentiles}. For more than half of the trips, the predictions of our model are off by less than 0.6\% and 0.4\% for fuel and electric consumption, respectively. The predictions on more than 90\% of the trips deviate from the ground truth by less than 5.7\% and 1.5\% for fuel and electric consumption, respectively.  
\begin{figure}[ht!]
\includegraphics[width=\linewidth]{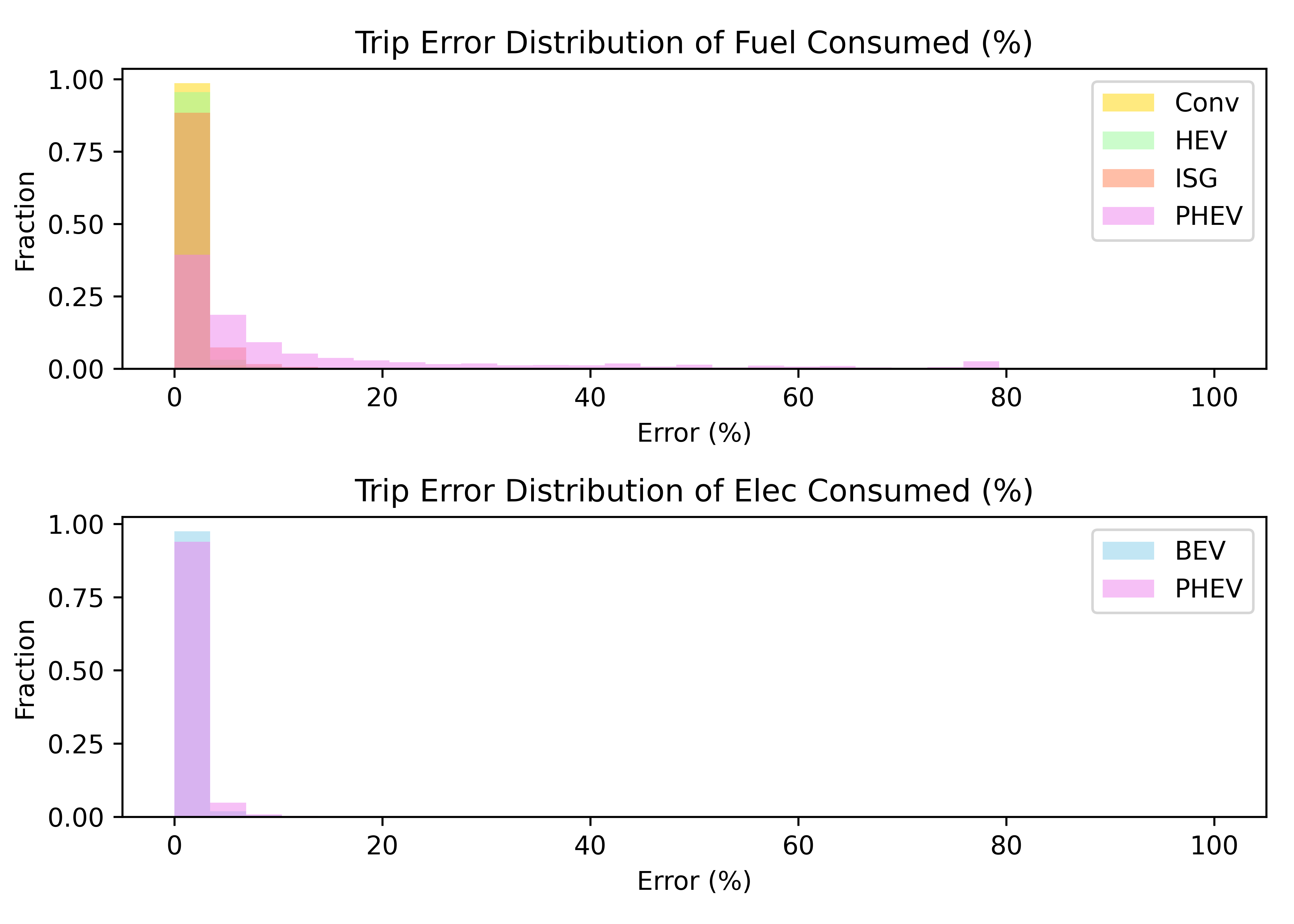}
\caption{Histogram of AAPE.}
\label{fig:HIST}
\end{figure}
Figure \ref{fig:HIST} shows the fractional histogram of AAPE. For conventional, HEV, and ISG powertrains, our model has relative errors less than 10\% for all but a negligible portion of trips. Our model also predicts most trips with APE errors less than 10\%, but it struggles on a small portion of trips. The lower sub-figure of Figure \ref{fig:HIST} shows that our model predicts all but a few trips with relative errors less than 5\%, which can also be seen in Table \ref{tab:percentiles}.

\begin{figure}[!t]
\centering
\subfloat[Electric consumption over links]{\includegraphics[width=\linewidth]{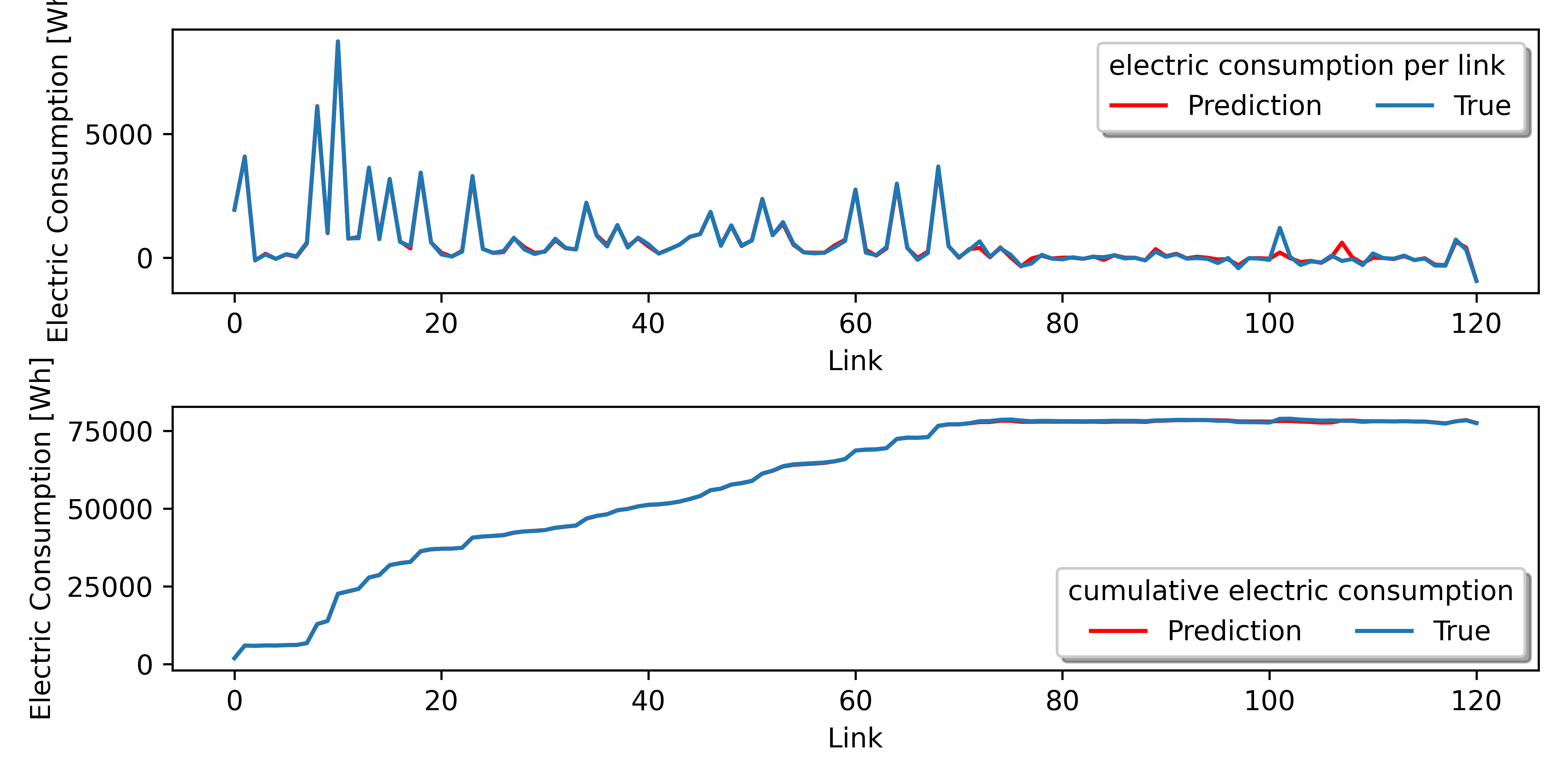}%
\label{fig:phev_ec}}
\\
\subfloat[Fuel consumed over links]{\includegraphics[width=\linewidth]{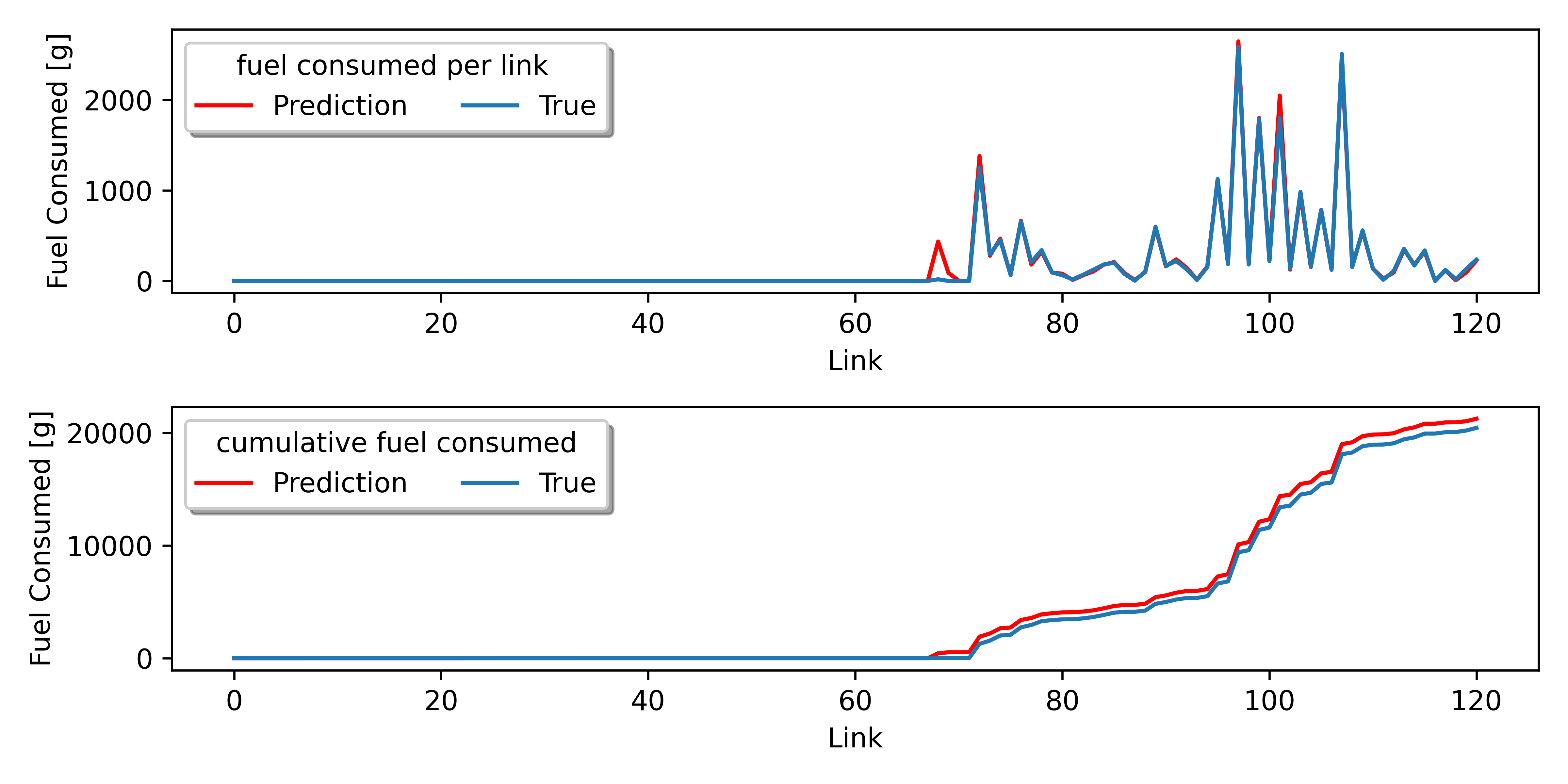}%
\label{fig:phev_fc}}
\caption{The energy consumption values of a PHEV on a trip. The model has trouble anticipating engine turn-on events at the charge depleting/charge sustaining transition.}
\label{fig:phev_trip}
\end{figure}

In Figure \ref{fig:phev_trip}, we see that a fully charged PHEV truck consumes electric energy at the beginning of a trip, as part of a charge depleting phase, and transitions to fuel consumption later in charge sustaining mode. The battery charge is depleted, and the model has some trouble anticipating an exact engine turn-on point. In other cases, due to the complexity of the PHEV engine controls, sporadic engine turn-on events are difficult to predict by this model. The indicator of such transitions or engine turn-on events is not completely contained in the features. As a result, the prediction of PHEVs is somewhat the least accurate. The relationship between macroscopic level vehicle speed and energy is, as expected, complex for PHEVs. This phenomenon is well studied in \cite{FIORI2019275}.

We summarize in Table \ref{tab:percentiles_type} the percentage error of the model broken down by powertrain. Overall, we conclude that the model does not show evidence of major bias towards short trips or towards specific powertrains and is adequate for subsequent tasks.

\begin{table}[]
\centering
\begin{tabular}{crrrr}
                                          & \multicolumn{4}{c}{\textbf{Error Quantiles (\%)}}                                                                 \\ \hline
\multicolumn{1}{c|}{\textbf{Veh Type}}    & \textbf{Mean}              & \textbf{50\%}              & \textbf{75\%}              & \textbf{95\%}              \\ \hline
\multicolumn{1}{c|}{\textbf{Conv}}        & {\color[HTML]{000000} 0.5} & {\color[HTML]{000000} 0.3} & {\color[HTML]{000000} 0.6} & {\color[HTML]{000000} 1.8} \\
\multicolumn{1}{c|}{\textbf{HEV}}         & 1.0                          & 0.5                        & 1.1                        & 3.2                        \\
\multicolumn{1}{c|}{\textbf{ISG}}         & 2.5                        & 0.8                        & 1.8                        & 6.3                        \\
\multicolumn{1}{c|}{\textbf{PHEV   fuel}} & 15.0                         & 5.0                          & 16.0                         & 69.0                         \\
\multicolumn{1}{c|}{\textbf{PHEV   elec}} & 0.7                        & 0.4                        & 0.7                        & 2.2                        \\
\multicolumn{1}{c|}{\textbf{BEV}}         & 1.2                        & 0.8                        & 1.5                        & 3.7                       
\end{tabular}
\caption{Percentage prediction error broken down by vehicle type}
\label{tab:percentiles_type}
\end{table}

\section{Recommendation System}
\subsection{TCO and Ranking System}
Our recommender system is based on the energy model trained in the previous section. It is assumed that a fleet usually comprises a limited number of trucks, perhaps less than two hundred. As a result, instead of using an approximate nearest neighbour search \cite{guo_accelerating_2020} to narrow down potential candidates into a small subset, we explicitly compute the estimated consumption of all vehicles for a given trip to give the best possible output. This brute force approach is more accurate and does not impose a computational burden given our setting. Let $V\subset \mathcal{R}^{D_2}$ denote the collection of all vehicles. Given a trip $\mathbf{U}\in \mathcal{R}^{\Tilde{T}\times D_1}$, we replicate $\mathbf{v}$ to match the length of the trip for each vehicle, concatenate the replicated tensor with route features to form input data point $\mathbf{X}$ for our model, and compute the estimated consumption for this vehicle, leveraging our model. The procedures of our recommendation system are shown in algorithm \ref{recsys}.

The next module extracts TCO values for the different candidates. The TCO is expressed in terms of \$/mile and is made up of two parts. First, a manufacturer's suggested retail price (MSRP) is calculated using a bottom up approach in which each component cost is determined based on technology and timeframe \cite{moawad_assessment_2016}, \cite{islam_extensive_2018}. MSRP is determined by summing up all component costs and multiplying by a retail price equivalent (RPE) factor of 1.2. It has been shown recently that such a rigid adjustment factor is not appropriate for all component technologies \cite{XAI_moawad} due to the large differences in pricing strategies between manufacturers, vehicle classes, market segments, etc. In the future we plan to leverage  top-down component cost estimates as presented in \cite{XAI_moawad} to improve MSRP estimation, the key to accurate TCO computation. 
The second part of the TCO calculation accounts for operational expenses. An energy cost is calculated based on the cost of electricity, gasoline, or diesel fuels using the \textit{Annual Energy Outlook} from the U.S Energy Information Administration (EIA) predictions \cite{EIA}. The energy cost is calculated over an expected period of ownership years and various vehicle miles traveled distributions that depend on the vehicle/application use. Other TCO components such as discount rates (inflation and risk premium), depreciation, and resale value are also considered in the calculation. Maintenance and insurance costs are not taken into account.

Based on TCO results, we designed a formula to rank the candidates on a scale of 1 to 5. Let $\mathbf{\Tilde{y}}\in \mathcal{R}^{n}$ be the estimated TCOs of $n$ candidates for a given trip. We rank each candidate by the following:
\begin{equation}
\mathbf{r}=\Big\lfloor4*\exp\Big({-\frac{\mathbf{\Tilde{y}}-\alpha}{\alpha}}\Big)+1\Big\rfloor
\label{eq:ranking}
\end{equation}
In equation \ref{eq:ranking}, we map $\frac{\mathbf{\Tilde{y}}-\alpha}{\alpha}$, the relative distance of TCOs to a target value $\alpha$, to set $\{1,2,3,4,5\}$. If we have some previous observations of energy consumption on the trip, we set $\alpha$ to be the expected optimal consumption, and let $\displaystyle \alpha = \min\{\alpha, \min_{1\leq i\leq n } \Tilde{y}_i\}$. If the expected optimal value is unknown, we set $\displaystyle \alpha = \min_{1\leq i\leq n } \Tilde{y}_i$.

\begin{figure}[ht!]
\includegraphics[width=\linewidth]{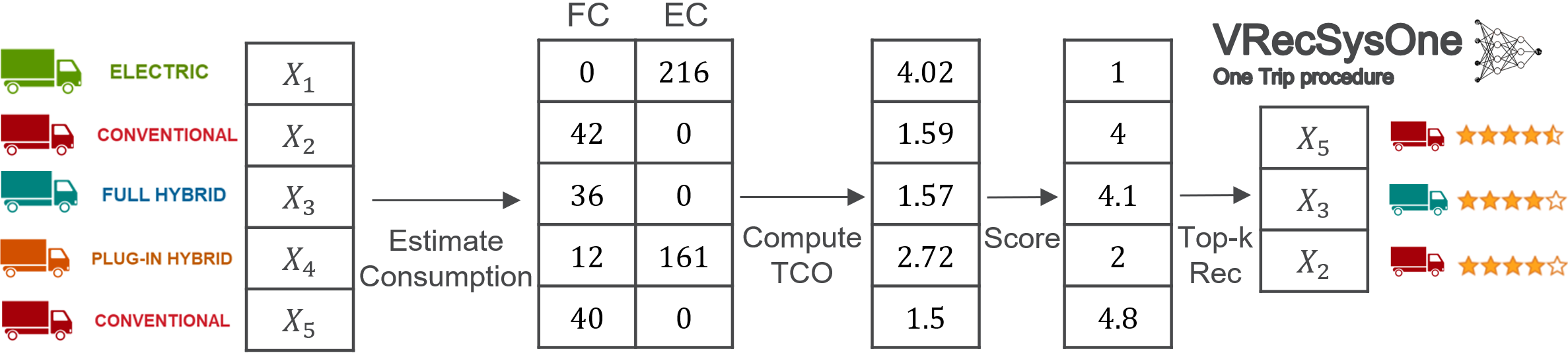}
\caption{Recommendation of five candidate vehicles evaluated for a single given trip. We call this framework VRecSysOne.}
\label{fig:VRecSysOne}
\end{figure}

\begin{algorithm}[ht!]
\SetKwInOut{Input}{input}\SetKwInOut{Output}{output}

\SetAlgoLined

\Input{A new trip $\mathbf{U}\in \mathcal{R}^{\Tilde{T}\times D_1}$, a set of candidate vehicles $V\subset \mathcal{R}^{D_2}$, evaluation model $f$.}
 \Output{Recommendation $\mathbf{v}^*$ and consumption table.}
 Table = \{\}\;
 \For{$\mathbf{v}\in$ $V$}{
  $\big([\mathbf{u}_1,\mathbf{v}], \cdots, [\mathbf{u}_{\Tilde{T}},\mathbf{v}]\big)\gets \mathbf{X}$\;
  $f(\mathbf{X})\gets \mathbf{y}$\ Estimate consumption\;
  $\mathbf{\Tilde{y}}\gets \mathbf{y}$ Compute TCO\;
 }
 $\mathbf{v}^*=\argmin{\mathbf{\Tilde{y}}}$
 \caption{Recommend vehicles from $V$}
 \label{recsys}
\end{algorithm}
Figure \ref{fig:VRecSysOne} shows an example of  five candidate vehicles recommended for a given trip.

\begin{figure}[ht!]
    \centering
    \includegraphics{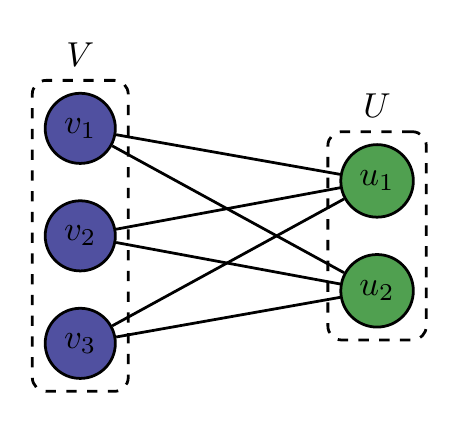}
    \caption{A weighted bipartite graph that represents candidates and trips. Each edge is weighted by TCO.}
    \label{fig:bipartite}
\end{figure}

\subsection{General Assignment Problem}
\begin{figure}[ht!]
\includegraphics[width=\linewidth]{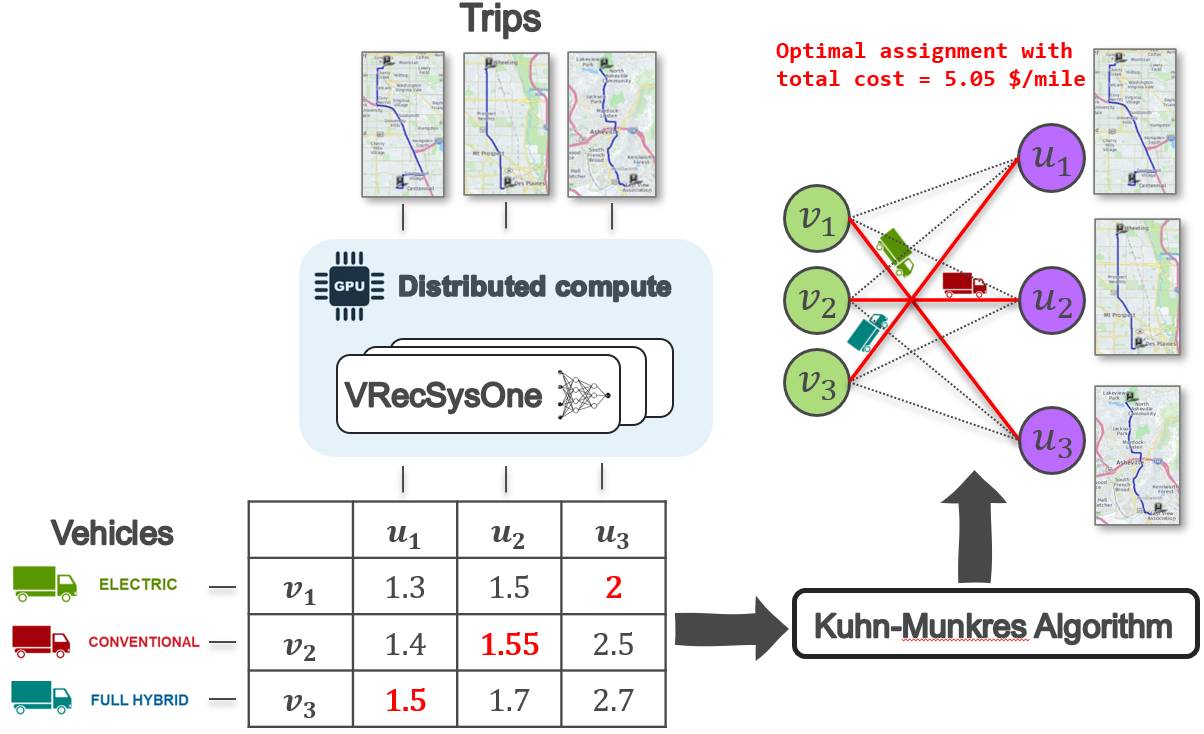}
\caption{Example  of a general assignment problem in which $n=m=3$, i.e., three trucks are to be optimally deployed over three trips such that fleet level TCO is minimized.}
\label{fig:assignment}
\end{figure}

In addition to computing TCO and performing single trip recommendations, we have extended the recommender system to optimize for a more general assignment problem. In this setting, given $n$ vehicles and $m$ trips ($n\geq m $), the goal is to select the best $m$ vehicles to minimize the total TCO of all $m$ trips. In other words, we would like to minimizes total TCO of the fleet in its entirety by optimally distributing/deploying trucks over various trips. As shown in Figure \ref{fig:bipartite}, this problem is equivalent to finding the minimum perfect matching of a complete weighted bipartite graph $G=(V \sqcup U, E)$. The algorithm that yields the optimal solution of this problem is the Kuhn-Munkres algorithm \cite{km}, which solves this problem via a primal-dual approach. The primal linear programming (LP) is formulated as
\begin{equation}
\begin{array}{ll@{}ll}
\text{minimize}  & \displaystyle\sum\limits_{i\in V, j \in U} w_{ij}&x_{ij} &\\
\text{subject to}& \displaystyle\sum\limits_{i \in V}   x_{ij} = 1,  & j\in U \\
&\displaystyle\sum\limits_{j \in U}   x_{ij} = 1,  & i\in V \\
&x_{ij}\geq 0, & i \in V, j\in U

\end{array}
\end{equation}
Thanks to the fundamental theorem of linear programming, the optimal solution of this problem will be found at a corner point over the feasible polytope. It guarantees the optimal solution is an 0-1 vector, i.e., $x_{ij}\in\{0,1\}, \forall i,j$. Hence, this LP is equivalent to our assignment problem. By solving the dual of the LP:
\begin{equation}
\begin{array}{ll@{}ll}
\text{maximum}  & \displaystyle\sum\limits_{i\in A, j \in B} u_i+v_j &\\
\text{subject to}& \displaystyle u_i+v_j\leq w_{ij},  & \forall i \in V, j\in U \\
\end{array}
\end{equation}
the K-M algorithm is guaranteed to obtain the optimal solution with complexity $\mathcal{O}(|V|\cdot |U|^2)$. The recommender system takes a $n$ by $m$ matrix consisting of TCOs and outputs a vector of length $m$ encoded IDs of optimal vehicles. Figure \ref{fig:assignment} shows an example of assignment input and output for the case of $n=m=3$. In this example, we note that the electric truck is not assigned to the first trip (best option) but instead assigned to the third trip (highest TCO). The full hybrid truck is deployed to the first trip, to benefit the total cost of the entire fleet.


\subsection{Application}
The recommendation system developed here is applied within a large-scale agent-based travel demand simulator called POLARIS \cite{auld2016polaris}. Applying such a model on hundreds of thousands of trips within a forecasting tool helps identify the possible spatial and temporal impacts of random and optimal vehicle type assignments for commercial trips. A brief overview of POLARIS and its components is  next, along with details on integration, and highlights of use case plans.
\subsubsection{POLARIS}
POLARIS is an agent-based activity-based travel demand forecasting tool that simulates all personal and freight travel for a given region across a 24-hour travel day. Residents in the region are synthesized based on underlying demographics to match all cross-tabulations of age, gender, household income, household size, and other key metrics from the U.S. Census Bureau's Public Use Microdata Sample (PUMS) \cite{pums}. 

On the personal travel side, activities are then planned and scheduled for traveler agents \cite{auld2012activity} from this synthetic population, and several econometric choice frameworks are applied to choose the destination for the activity, start time, and travel mode to travel to the chosen activity location. 

Currently, on the freight side, commercial trips both inside and external to the region (i.e., trips originating or terminating outside the region) are modeled through a top-down framework that decomposes aggregate data into individualized trips so that congestion patterns can be calibrated for a region \cite{noauthor_smart_nodate}. A full agent-based freight model was first implemented in \cite{stinson_large-scale_2020} including an e-commerce MDT module\cite{stinson2019}, and is being extended with data-driven processes and a robust framework named CRISTAL (Collaborative, Informed Strategic Trade Agents with Logistics) as proposed in \cite{stinson_introducing_2021}. The CRISTAL framework creates a synthetic population of firm agents and their member establishments using data from the US Census County Business Patterns and other sources. Fleets for each firm are modeled using a Seemingly Unrelated Regression with Latent Tobit Variables (SURTLV) approach \cite{stinson_method_2020} that includes variables for customer service and logistics sophistication strategies of firms \cite{inproceedings} using data from the CoStar real estate database, the US Securities and Exchange Commission, and other sources. The SURTLV model predicts the number of MDT and HDT owned by each firm. The CRISTAL framework is being integrated with POLARIS to create a fully agent-based passenger and freight model that supports several travel modes ranging from household vehicles (single or high-occupancy trip), non-motorized travel (walk or bike), use of a ride-hailing fleet \cite{gurumurthy2020integrating}, and medium and heavy-duty freight trips. With all travel demand in place, a time-dependent $A^*$ router provides trajectories taking into account travelers' expected user equilibrium path through information mixing from historical information \cite{auld2019agent}. Vehicles move on the network abiding by traffic flow principles \cite{de2019mesoscopic} with powertrain-specific decision-making available such as electric vehicle battery consumption and charging.

\subsubsection{Model Integration and plans}
POLARIS is a powerful large-scale travel demand model encompassing several travel behaviors for a realistic forecast. Integrating the recommendation system into POLARIS was done through the Lite interface of the large-scale machine-learning estimation and prediction tool TensorFlow \cite{tensorflow2015-whitepaper}. TensorFlow Lite provides a lightweight interface for machine-learning model prediction and is typically used to deploy models on micro devices, such as phones and tablets. With its low latency and low memory framework, the recommendation system can be used within POLARIS with minimal impact on memory or computational footprint while still leveraging robust results.

The next steps will be first to integrate the recommendation system with each fleet, then to deploy the integrated system in studies involving alternative powertrains for MD/HD freight, with powertrain options including diesel, gasoline, electric and hydrogen. The system will be implemented at the operational level, where routes are generated for each vehicle that is conducting pickup and delivery tours. Information on these operational demands (the routes and their characteristics such as length) will be used with a feedback loop to the recommender system to evaluate the optimal powertrain mix for the fleet. The resulting route and powertrain outputs will be combined to evaluate energy and emissions for each fleet, which can be aggregated up to study regional freight system impacts.

\section{Conclusion and Discussion}
This work shows how an end-to-end vehicle\textendash route recommender system framework can be designed by leveraging large-scale trip/route data and deep learning models. In our setup, the trip, route, vehicle and energy data are generated by a suite of transportation system and vehicle modeling tools using a wide range of representative mobility scenarios across an entire metropolitan area in varying conditions. We presented a latent energy learning model that attempts to correlate energy outcomes at the link (road segment) level with road and high-level trip information only, i.e., with masked high-resolution speed profiles. On the basis of highly accurate energy models, subsequent blocks of the algorithm perform single trip recommendations for a list of vehicle candidates by considering TCO metrics. In its more general form, we have implemented a general assignment functionality for which the recommender system can be leveraged to optimally deploy several trucks over several defined trips. For fleets, this can support investment decisions (e.g., which combination of vehicle technologies would allow to minimize TCO) as well as match individual vehicles to routes on any particular day. It represents real world situations in which routing can be known up to a certain level of traffic fidelity. With a given truck inventory and/or availability specific to the  fleet, the recommender model facilitates cost-effective deployment of trucks to routes. Finally, we've deployed and integrated the model into the POLARIS agent-based transportation system tool for online and real-time truck recommendations within the modeled fleets.

Route selection has a big impact on the vehicle technology choice. Different powertrain choices, and investment decisions are made on the basis of cost, energy, time or other psychological aspects. For example, \cite{FIORI2018262} investigates this optimum routing assignment for electric vehicles (in comparison to conventional vehicles) via specific energy modeling and microscopic simulation (i.e with known speed profile). \cite{NIE2013154} and \cite{6422511} studied eco-routing comparisons between various vehicle technologies while retaining micro information. \cite{AHN2008151}, among others \cite{NESAMANI20072996}, analyze thoroughly conditions for energy efficient assignment, leveraging data and simulation, and point out that macroscopic tools lack of fidelity for an accurate energy estimation. In fact, \cite{FONTES2015293} discusses the complexity involved. In our work, we overcome the need to depend on microscopic level information, this was tackled in earlier studies via different approaches \cite{atmos12010053}, \cite{ZEGEYE2013158}, \cite{RODRIGUEZREY2021102725}, \cite{OSORIO2015520}, \cite{SUN201527}, and \cite{6236175} developed a sophisticated eco-routing framework on that basis. We proposed here a deep learning perspective to predict accurate energy consumption values in a macroscopic setting, that learns latent (microscopic) information via careful model design and feature engineering. Under the umbrella of a recommender system, we applied the framework to routing, and technology dependent cost-efficient assignment problems as attempted previously by \cite{JIMENEZ2016120}. This new framework is very relevant, for example, to freight companies.

\section*{Acknowledgment}
The work described was sponsored by the U.S. Department of Energy (DOE) Vehicle Technologies Office (VTO) under the Systems and Modeling for Accelerated Research in Transportation (SMART) Mobility Laboratory Consortium, an initiative of the Energy Efficient Mobility Systems (EEMS) Program. The following DOE Office of Energy Efficiency and Renewable Energy (EERE) managers played important roles in establishing the project concept, advancing implementation, and providing ongoing guidance: Erin Boyd and Danielle Chou (Office of Energy Policy and Systems Analysis, U.S. Department of Energy). The submitted manuscript has been created by UChicago Argonne, LLC, Operator of Argonne National Laboratory (Argonne). Argonne, a U.S. Department of Energy Office of Science laboratory, is operated under Contract No. DE- AC02-06CH11357. The U.S. Government retains for itself, and others acting on its behalf, a paid-up nonexclusive, irrevocable worldwide license in said article to reproduce, prepare derivative works, distribute copies to the public, and perform publicly and display publicly, by or on behalf of the Government.\par

\bibliography{bibtex/IEEEabrv.bib,bibtex/reference.bib}{}
\bibliographystyle{IEEEtran}

\begin{IEEEbiographynophoto}{Ayman Moawad}
is a research engineer in the Vehicle and Mobility Simulation group at Argonne National Laboratory. He received a master's degree in Mechatronics, Robotics, and Computer Science from the Ecole des Mines, France and a master's degree in Statistics from the University of Chicago, USA. His research interests include engineering applications of artificial intelligence for energy consumption and cost prediction of advanced vehicles, machine learning, large scale data analysis, and high performance computing.
\end{IEEEbiographynophoto}

\begin{IEEEbiographynophoto}{Zhijian Li}
is a Mathematics PhD student at the University of California, Irvine. He received his bachelor's degree in Applied Mathematics from University of California, Los Angeles. His current research interest lies in compression of neural networks, including quantization-aware training, sparsification, and channel pruning. He has also worked on spatio-temporal time-series forecasting through graph-based recurrent neural networks.
\end{IEEEbiographynophoto}

\begin{IEEEbiographynophoto}{Ines Pancorbo}
is currently a Data Scientist at Visa and was previously a research aide in the Vehicle and Mobility Simulation group at Argonne National Laboratory. She received a master’s degree in Mathematics and Statistics from Georgetown University, USA, and a bachelor’s degree in Mathematics from the University of Maryland, USA. She is interested in applications of machine learning and deep learning for prediction purposes as well as causal inference.
\end{IEEEbiographynophoto}

\begin{IEEEbiographynophoto}{Krishna Murthy Gurumurthy}
is a Computational Transportation Engineer in the Transportation Research Systems Modeling and Control Group at Argonne National Laboratory. He received his doctorate for research focusing on travel demand modeling \& forecasting, especially through the utilization of large-scale agent-based simulation tools. He is particularly interested in capturing the impacts of shared and automated vehicles on travel patterns and congestion and measuring the resulting effects of various policies.
\end{IEEEbiographynophoto}

\begin{IEEEbiographynophoto}{Vincent Freyermuth}
is a research engineer in the Transportation Systems Simulation and Vehicle and Mobility Simulation Group at Argonne National Laboratory. Vincent has 20 years of experience in the area of vehicle simulation, vehicle integration and product development. He links agent-based transportation level models (POLARIS) to detailed vehicle level models (Autonomie) and a life cycle GHG analysis tool (GREET) to understand the broader impact of emerging mobility trends.
\end{IEEEbiographynophoto}

\begin{IEEEbiographynophoto}{Ehsan Islam}
completed his M. Sc. in Interdisciplinary Engineering from Purdue University, USA in 2019 and B.A.Sc in Mechatronics Engineering from University of Waterloo, Canada in 2016. His focus is on applying mechatronics principles to innovate processes in advanced vehicle technologies and controls systems. At Argonne, he focuses his research on vehicle energy consumption analyses and inputs for U.S. DOE-VTO and NHTSA/EPA/U.S. DOT CAFE and CO$_{2}$ standards using innovative large scale simulation processes and applications of AI.
\end{IEEEbiographynophoto}

\begin{IEEEbiographynophoto}{Ram Vijayagopal}
is the technical manager for Vehicle Technology Assessment at Argonne National Laboratory. He is responsible for quantifying the energy saving potential of technologies using modeling and simulation. After working at Mahindra \& Mahindra and Hitachi Automotive Systems, he joined Argonne in 2008. He received his bachelor’s degree in engineering from University of Kerala  and a master’s degree in engineering from University of Michigan.
\end{IEEEbiographynophoto}

\begin{IEEEbiographynophoto}{Monique Stinson}
is Technical Manager, Freight Systems and Analytics in Argonne National Laboratory’s Vehicle and Mobility Systems Section and lead developer of CRISTAL (Collaborative, Informed, Strategic Trade Agents with Logistics) that is deployed in Argonne’s flagship POLARIS agent-based software (large-scale transportation system simulation).  Dr. Stinson has researched the energy, emissions, mobility, productivity and equity impacts of freight transportation systems in Chicago, Detroit, Atlanta, Phoenix, California, and other areas for nearly 20 years. Her research spans the system-level effects of vehicle technologies, connectivity and automation, e-commerce, commodity flow growth, and other scenarios, to support stakeholders ranging from DOE to industry to local governments. Dr. Stinson is a member of the American Transportation Research Institute (ATRI) Research Advisory Committee (RAC), Transportation Research Board’s (TRB) Freight Planning and Logistics Committee, and the TRB Freight Data Committee, and is a contributor to the 21st Century Truck Partnership (21CTP) Freight Operational Efficiency Tech Team. 
\end{IEEEbiographynophoto}

\begin{IEEEbiographynophoto}{Aymeric Rousseau}
is the Manager of the Vehicle and Mobility Simulation group at Argonne National Laboratory. He received his engineering diploma at the Industrial System Engineering School, France in 1997 and an Executive MBA from Chicago Booth in 2019. For the past 20 years, he has been evaluating the impact of advanced vehicle and transportation technologies from a mobility and energy point of view, including the development of Autonomie (vehicle system simulation) and POLARIS (large-scale transportation system simulation).
\end{IEEEbiographynophoto}

\end{document}